\documentclass[]{fairmeta}

\usepackage[utf8]{inputenc}
\usepackage{array}
\usepackage{amsfonts}
\usepackage{amsmath}
\usepackage{amssymb}
\usepackage{amsthm}
\usepackage{nicefrac}
\usepackage{enumitem}
\usepackage{algorithm}
\usepackage{algorithmic}
\usepackage{flafter}
\usepackage{tikz}

\graphicspath{{figures/}}
\hypersetup{
  pdftitle={Quantizing Looped Transformers: Feedback Exposure and Calibration Blindness},
  pdfauthor={Nux Li},
  pdfsubject={Post-training quantization for looped transformers}
}

\title{Quantizing Looped Transformers:\\Feedback Exposure and Calibration Blindness}

\author[1]{Nux Li}
\affiliation[1]{Meta}

\abstract{Looped transformers reuse weights across recurrence steps, making low-bit quantization
especially attractive. We identify two distinct failure modes of standard post-training
quantization. On Huginn-3.5B, per-channel INT4 fails primarily at the non-residual loop-entry
adapter, while quantizing the residual core is much less damaging. We call this
\emph{feedback exposure}: a quantized layer perturbs the recurrent state without an identity path,
and the resulting error is fed back at later steps. Controlled experiments on linear filters and
Mamba state-space models show that feedback exposure also occurs outside transformers. Grouped
INT4 reveals a separate failure, \emph{calibration blindness}: our one-step GPTQ baseline
builds its Hessian from step-0 activations, leaving input directions used later in the recurrence
nearly unweighted. Across nine checkpoints from seven looped architectures,
one-step GPTQ is worse than round-to-nearest (RTN) on the primary task metric for five
checkpoints. Accumulating the GPTQ Hessian across recurrence steps outperforms both one-step GPTQ and
RTN on all nine checkpoints and recovers bf16-level accuracy on Huginn. These results
separate two questions for PTQ on looped models: where quantization error enters the recurrence,
and which states calibration sees.}

\date{September 2026}
\correspondence{Nux Li at \email{nux@meta.com}}

\begin{document}

\maketitle

% "Quantizing Looped Transformers: Feedback Exposure and Calibration Blindness"
% arXiv preprint

\section{Introduction}

Looped transformers iterate a shared layer block $N$ times, decoupling computational depth from parameter count. Huginn~\citep{geiping2025huginn} has 3.5B parameters but unfolds to an effective depth of 132 layers at 32 recurrences; Ouro-1.4B and Ouro-2.6B~\citep{ouro2025} reuse 24- and 48-layer stacks for four loop steps (96 and 192 block applications), and COCONUT~\citep{hao2024coconut} recycles GPT-2's entire forward pass. For single-pass transformers, INT4 quantization roughly quarters the weight footprint. We test whether that win transfers to looped models where the same weights are reused across recurrence steps.

On a 100-example GSM8K diagnostic, per-channel INT4 on the loop-entry adapter and recurrent core reduces exact match
from 34\% in bf16 to 0\% (Appendix Table~\ref{tab:perchannel-diagnostic}). To localize this
collapse, Figure~\ref{fig:hero}a compares each configuration's token predictions with bf16
on a separate 50-prompt diagnostic.
Adapter-only INT4 ends at 10.5\% agreement, nearly matching 10.6\% when both components
are quantized; core-only INT4 retains 87.9\%. COCONUT's residual GPT-2 stack behaves
differently, with agreement rising from 44.8\% after the base forward to 61.5\% after 8 latent
recurrences (Figure~\ref{fig:hero}b).

Standard one-step sensitivity metrics do not explain this split. Small random input perturbations produce the least local output change at the adapter (17th of 17 projections), while Hessian trace ranks a core projection as most sensitive even though quantizing that projection alone at INT4 group size 128 changes far fewer predictions than quantizing the adapter. These metrics omit where the rounding error enters the recurrence. Section~\ref{sec:mechanism} isolates this mechanism; linear-filter and Mamba controls show that transition perturbations are also more damaging outside transformers.

Grouped INT4 ($g{=}128$) avoids the immediate collapse, but one-step GPTQ still underperforms round-to-nearest (RTN) on five of nine checkpoints. Its step-0 Hessian misses directions used later in the recurrence, a mismatch that persists with more calibration prompts. These mechanisms are independent. Feedback exposure identifies where perturbations are structurally dangerous, while calibration blindness identifies when GPTQ calibrates the wrong subspace. Huginn's adapter lies at their intersection. We isolate the first with controlled interventions and address the second with trajectory calibration across nine checkpoints from seven looped architectures.

% Section 2 — Background
% arXiv preprint

\section{Background}
\label{sec:background}

\paragraph{Looped transformers.}
A looped transformer applies a shared block iteratively to a latent state:
\begin{equation}
  s_{t+1} = R_\theta(s_t, e, c_t), \qquad t = 0, \ldots, N{-}1,
  \label{eq:recurrence}
\end{equation}
where $s_t$ is the latent state, $e$ is the fixed prompt embedding, and $c_t$ is architecture-specific conditioning.

Huginn-3.5B~\citep{geiping2025huginn} and Recurrent-Llama-1.4B (R-Llama)~\citep{mcleish2025rllama} route recurrence through a non-residual adapter that projects the concatenation of $s_t$ and $e$ to $d$ dimensions; their recurrent cores remain residual. Other looped models reuse residual layers without a dedicated adapter~\citep{ouro2025,jeddi2026loopformer,hao2024coconut} or impose explicit stability constraints on the recurrent core~\citep{prairie2026parcae}.

We use the local spectral radius $\rho$ of the per-step Jacobian $\partial s_{t+1}/\partial s_t$ as a first-order stability diagnostic; $\rho > 1$ indicates linearized amplification in some direction. The echo state property requires trajectories driven by the same input to forget their initial conditions~\citep{jaeger2004harnessing}; uniform contractivity is a sufficient global condition, whereas our per-step $\rho$ is a local surrogate. For GPTQ calibration, we report the rank of $H=X^\top X$ at threshold $10^{-3}\lambda_{\max}$, written $\#\{\lambda_i(H) > 10^{-3}\lambda_{\max}\}$, which counts calibration directions above a fixed relative energy threshold.

\paragraph{Post-training quantization.}
Per-channel INT4 quantizes each output row with a single scale factor. At $g{=}128$, each group of 128 consecutive columns gets its own scale, which lowers quantization error but adds storage overhead. GPTQ~\citep{frantar2022gptq} rounds weights column by column to minimize $\|XW^\top - X\hat{W}^\top\|_F^2$, where each row of $X$ is a calibration activation and $H = X^\top X$ is the per-layer Hessian proxy. To spread outlier energy, QuIP\#~\citep{chee2024quip} and SpinQuant~\citep{ashkboos2024spinquant} apply orthogonal transforms across dimensions. QuIP\# adds lattice codebooks for sub-4-bit targets. AWQ~\citep{lin2024awq} instead rescales channels by activation magnitude and does not use a Hessian proxy. To test whether step-0 Hessians miss deployed directions, we compare GPTQ using $H$ from each shared layer's first invocation with GPTQ using $H$ accumulated over all $N$ invocations.

\paragraph{Quantization under iteration.}
Q-Diffusion~\citep{qdiffusion2023} samples calibration data across denoising timesteps, while AdaTSQ~\citep{adatsq2024} combines timestep-dependent activation policies with temporally weighted calibration for static weights. Both address activation shifts along recursively generated denoising trajectories. Looped transformers create a related shift by feeding the shared block's output back as its next latent state. For linear time-invariant systems, Quamba~\citep{quamba2024} bounds quantized error when the transition matrix $A$ is unperturbed. \citet{li2026mondeq} analyze quantized monotone DEQs and show that convergence survives operator perturbation when the monotonicity margin is large enough. The checkpoints studied here come with no such certificate.

\paragraph{Residual robustness under reuse.}
\citet{veit2016residual} showed that ResNets tolerate layer \emph{deletion} at test time. Their path-ensemble argument implies that deleting one of $L$ residual blocks removes $2^{L-1}$ of the $2^L$ paths. The argument is feedforward because each residual block appears at most once on any path. In a looped model, the same weight perturbation acts on the recurrent state at every step, so this path-counting argument does not describe how its effect evolves across recurrence.

% Section 3 — Diagnosis
% arXiv preprint

\section{Feedback Exposure Under Recurrence}
\label{sec:mechanism}

\begin{figure}[!t]
\centering
\includegraphics[width=0.96\linewidth]{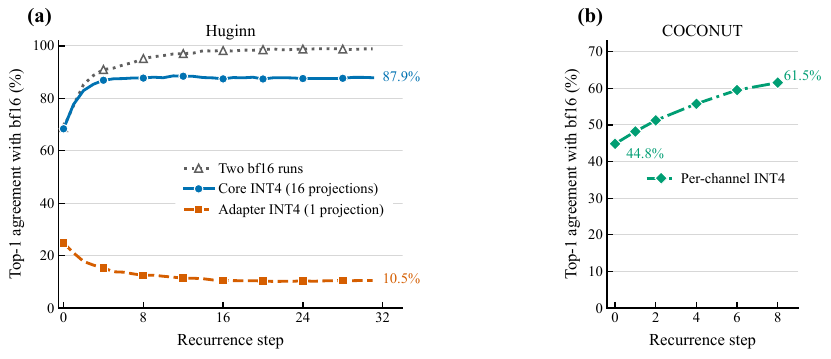}
\caption{Per-channel INT4 feedback exposure. (a)~Huginn's non-residual adapter-only agreement falls to 10.5\%, while quantizing all 16 residual-core projections together retains 87.9\%. Two bf16 runs with independently sampled initial states reach 98.9\% agreement. (b)~On 15 fixed prompts, COCONUT agreement rises from 44.8\% after the quantized base forward to 61.5\% after 8 latent recurrences.}
\label{fig:hero}
\end{figure}

A looped transformer applies the same quantized weights at every step of the recurrence, but not all shared layers contribute equally to the resulting damage. We separately quantize the adapter and core with per-channel INT4 in Huginn and R-Llama to isolate which layers are responsible. Grouped INT4 ($g{=}128$) is used for the sensitivity-ranking comparison in Appendix~\ref{sec:app-sensitivity} and for the recovery experiments in Sections~\ref{sec:method} and~\ref{sec:experiments}.

Unless noted, the Huginn and R-Llama diagnostics use deterministic symmetric
round-to-nearest per-channel INT4 with simulated float32 dequantization, with all other
layers in bf16. Agreement is the fraction of token
positions whose top-1 prediction matches bf16 at the same step, averaged equally over
the first 50 GSM8K test prompts.

\subsection{Controlled Injection}
\label{sec:controlled}

We first isolate the adapter and recurrent core. Adapter-only INT4 drops final-step
agreement with bf16 to 10.5\%, whereas quantizing all 16 core projections leaves 87.9\%
(Figure~\ref{fig:hero}a). Agreement under core-only INT4 rises over the recurrence, and
quantizing both components yields 10.6\%, matching the adapter-only result.

To test whether recirculating prior state error drives this damage, we vary how much of
that error reaches the adapter at the next step. This control uses 50 GSM8K training
prompts disjoint from calibration and 3 recurrent-state initializations shared across all
$\lambda$ values. Writing $s_t^\lambda$ for the intervened
state, the quantized adapter receives
$s_t^{\mathrm{bf16}}+\lambda(s_t^\lambda-s_t^{\mathrm{bf16}})$, where $\lambda=0$ removes
prior state error and $\lambda=1$ gives the ordinary quantized rollout. Removing prior
error raises final-step agreement from 10.4\% to 69.2\%. The relative final-state error
$\|s_{32}^\lambda-s_{32}^{\mathrm{bf16}}\|_2/\|s_{32}^{\mathrm{bf16}}\|_2$ grows
monotonically with $\lambda$ under both per-channel and grouped INT4. Quantizing the
recurrent-state and fixed-embedding columns separately gives the same trend, although
the feedback effect is larger for the embedding columns (Appendix~\ref{sec:app-arms}).

R-Llama-1.4B shows the same asymmetry at smaller scale: adapter-only agreement is 49.1\%,
compared with 74.8\% for core-only (Appendix Table~\ref{tab:controlled-rllama}). Standard
sensitivity rankings also miss this asymmetry. In Huginn's grouped-INT4 single-projection ablation,
the adapter causes the most damage, although small random input perturbations rank it last
and Hessian trace ranks it third (Appendix
Table~\ref{tab:sensitivity-ranking}).

The adapter/core asymmetry persists under other weight perturbations (Appendix~\ref{sec:app-perturbation-types}). At the same Frobenius norm for both locations, a random rank-4 weight update leaves 71.8\% agreement at the adapter and 95.8\% at the core, a 24.0\,pp gap. A Gaussian weight perturbation scaled separately at each layer to match its grouped-INT4 output-error norm likewise leaves lower agreement at the adapter than at the core (89.3\% vs.\ 96.5\%), despite a smaller matched norm at the adapter (3.16 vs.\ 8.11). By comparison, per-channel INT4 rounding at the same adapter position drops agreement to 10.5\%.

Together, these controls tie the collapse to the error's position in the recurrence,
its repeated feedback, and the structure of the perturbation.

By contrast, COCONUT reuses GPT-2's full transformer stack across latent steps and has no dedicated adapter, so every layer sits on an identity path. Under per-channel INT4 on the shared stack, agreement with bf16 improves from 44.8\% after the quantized base forward ($N{=}0$ latent recurrences) to 61.5\% after 8 latent recurrences, and KL divergence falls from 1.25 to 0.93 (Figure~\ref{fig:hero}b).

\subsection{Feedback Exposure and the Identity Path}
\label{sec:residual}

At a fixed operating point, the Jacobian of one full recurrence step with respect to
$s_t$ factors as
\begin{equation}
  J = \underbrace{\textstyle\prod_{i}\bigl(I + \tfrac{\partial F_i}{\partial x}\bigr)}_{J_{\text{core}}} \;\times\; W_s,
  \label{eq:jacobian}
\end{equation}
where $F_i$ is the net residual update of core block $i$. Writing the linear adapter as $W_s s_t + W_e e + b$, quantization changes its output by $\Delta W_s s_t + \Delta W_e e$. The first term directly perturbs the state-input Jacobian, while both errors enter the next recurrent state through the same unprotected adapter output. No identity path carries $s_t$ around the adapter. Each core factor includes the identity ($I + \partial F_i/\partial x$, with RMSNorm, attention, and MLP nonlinearities inside $F_i$), so each block retains an additive identity path for its input. We call a perturbation \emph{feedback-exposed} when it acts on the recurrent state without an identity path around it.

\paragraph{Spectral radius.} For each prompt, we evaluate the bf16 and INT4 Jacobians at its bf16 state after 32 recurrence steps. The mean $\rho$ is 1.064 for bf16 and 1.145 after per-channel adapter INT4; the INT4 map has $\rho>1$ on all 30 prompts, versus 9/30 for bf16.

\paragraph{Identity-bypass control.} To isolate the bypass, we hold each prompt's state and the bf16/INT4 adapter weights fixed, then apply the same small state perturbations with and without adding $s$ around the adapter. The bypass reduces the RMS difference between the bf16 and INT4 one-step changes by 37.4\% (41.7\% after normalizing each difference by its bf16 response magnitude; Appendix~\ref{sec:app-jacobian}).

R-Llama fails differently from Huginn. Per-channel RTN reduces GSM8K accuracy to 0\%, but the measured trajectory displacement stabilizes quickly. Starting from the bf16 state after 32 recurrence steps, a diagnostic rollout advances the bf16 and quantized trajectories for 32 more updates and measures $\|s_t^{\mathrm{q}}-s_t^{\mathrm{bf16}}\|_2$. For both per-channel and $g{=}128$ INT4, mean displacement is within 1\% of its final value after 4 updates. The final norm is 1082 under per-channel INT4 and 300 at $g{=}128$ (Appendix~\ref{sec:app-error-amp}). This finite-horizon saturation is consistent with a displaced fixed point: the measured trajectory remains bounded without returning to the bf16 state.

\paragraph{Feedback exposure in IIR filters.} An infinite-impulse-response (IIR) filter
isolates the same structural distinction in a linear recurrence. Its matrix $A$ feeds the
previous state into the next step, while $B$ injects the current input and $C,D$ produce the
output. The poles are the eigenvalues of $A$, so only quantizing the feedback transition can
change stability. Quantizing $A$ moves at least one pole outside the unit circle in all 4
tested Butterworth, Chebyshev, and elliptic filters at 10 bits or below; quantizing $B,C,D$
leaves every pole unchanged across all 6 tested precisions (Appendix~\ref{sec:app-iir}).

\paragraph{Feedback exposure in SSMs.} Mamba~\citep{gu2023mamba} shows the same distinction between perturbing the state transition and perturbing components outside it. On Mamba-2.8B, let $d_t=\|h_t^{\mathrm{INT4}}-h_t^{\mathrm{bf16}}\|_2/\|h_t^{\mathrm{bf16}}\|_2$ denote the relative SSM-state error at token $t$. Under symmetric INT4 round-to-nearest on \texttt{A\_log}, the ratio of median $d_{32}$ to median $d_8$ ranges from $1.30$ to $1.71$ across 5 perturbed layers and 10 prompts. Perturbing \texttt{out\_proj} changes the block output but leaves the same layer's SSM state unchanged because the projection follows the scan. The recurrence explains this behavior: $A=-\exp(A_{\log})$ and $\bar{A}_t=\exp(\Delta_t A)$, so quantizing \texttt{A\_log} changes the multiplier on $h_{t-1}$ at every token even though each diagonal entry of $\bar{A}_t$ lies in $(0,1)$. In a Vim-S 4-bit activation ablation, QMamba~\citep{qmamba2024} reports a 66.8-point top-1 drop from quantizing $\bar{A}_t$, the largest among its tested SSM activations, while Bi-Mamba~\citep{bimamba2024} retains the SSM parameters at full precision and binarizes the projections. Both results are consistent with feedback exposure (Appendix~\ref{sec:app-mamba}).

\paragraph{Mitigating feedback-exposed layers.}
In Huginn at $g{=}128$, the adapter is 1 of 17 shared projections. In a 100-example GSM8K ablation, keeping only the adapter in bf16 raises exact-match accuracy from 25\% to 32\%, recovering 7 of the 9 points lost to quantization (Appendix Table~\ref{tab:ablation}). At per-channel INT4, adapter-only Hadamard rotation with trajectory calibration reaches 29.4\% exact match on all 1{,}319 GSM8K test examples, compared with 34.9\% in bf16 (Appendix~\ref{sec:app-rotation-type}). Huginn's localized feedback path permits these projection-specific fixes. Ouro and LoopFormer distribute recurrence across the full stack, so we evaluate trajectory calibration instead (Section~\ref{sec:calibration}).

% Section 4 — Calibration Blindness
% arXiv preprint

\section{Calibration Blindness from Step-0 Hessians}
\label{sec:calibration}

One-step GPTQ underperforms round-to-nearest on five of nine checkpoints (Table~\ref{tab:main}). This grouped-INT4 failure has a different cause from the per-channel collapse in Section~\ref{sec:mechanism}. The Hessian solver can optimize the wrong subspace even when immediate collapse is absent. Our one-step GPTQ baseline rounds weights to minimize a Hessian-weighted proxy $\mathrm{tr}(\Delta W\, H_0\, \Delta W^\top)$, where $H_0$ is built from step-0 activations alone. For layers whose input shifts across recurrence, the step-0 proxy leaves most directions weakly penalized.

\begin{table}[H]
\centering
\caption{Per-layer thresholded Hessian rank at $10^{-3}\lambda_{\max}$ on 50 GSM8K prompts for Huginn-3.5B. The adapter's step-0 Hessian covers only 43 of 10{,}560 input dimensions. Accumulating 32 recurrence steps raises this to 335.}
\label{tab:rank}
\small
\begin{tabular}{@{}lrrrrr@{}}
\toprule
\textbf{Layer} & \textbf{Dimension} & \textbf{1-step rank} & \textbf{1-step \%} & \textbf{32-step rank} & \textbf{Gain} \\
\midrule
\textbf{Adapter} & 10{,}560 & \textbf{43} & \textbf{0.41\%} & \textbf{335} & \textbf{7.8$\times$} \\
Core Wqkv (avg) & 5{,}280 & $\sim$1{,}050 & 20\% & $\sim$1{,}450 & 1.3$\times$ \\
\texttt{mlp.proj} (block~3) & 17{,}920 & 65 & 0.36\% & 89 & 1.4$\times$ \\
\bottomrule
\end{tabular}
\end{table}

Of 10{,}560 adapter input dimensions, only 43 exceed the $10^{-3}\lambda_{\max}$ threshold in $H_0$. Greedy column compensation can reduce the step-0 proxy while leaving error along directions that the proxy weights weakly. Later recurrence steps activate additional directions, so a small step-0 proxy does not guarantee low error along the deployed trajectory. Core Wqkv starts at 20\% coverage and gains only $1.3\times$ when later-step activations are included, compared with $7.8\times$ for the adapter.

\paragraph{1{,}000-prompt control.} At step~0, the adapter concatenates a noise-initialized recurrent state $h$ with the content-dependent prompt embedding $e$. Across 1{,}000 prompts, these inputs still occupy a narrow step-0 subspace. To compare sample counts, we use the row-averaged Hessian $\widetilde H=2X^\top X/m$ for $m$ activation rows; this scaling leaves thresholded rank unchanged. Rank falls from 43 at 50 prompts to 20 at 1{,}000 while the top eigenvalue changes by less than 0.3\%, as near-threshold eigenvalues shrink with more samples. The top-32 eigenspaces of the uncentered recurrent-state second moments at steps~0 and~31 are nearly orthogonal across 2 initialization seeds (mean squared cosine of principal angles: 0.57--0.63\%; Appendix~\ref{sec:app-eigenvalue}), confirming that step-0 data does not cover later-state directions.

\paragraph{Per-step rank.} Holding the same 2{,}919 prompt-token positions across steps, Huginn's rank rises from 43 at step~0 to 259 at step~1 and 363 at step~4, ruling out sample accumulation. Ouro and LoopFormer also show large first-step jumps, while COCONUT changes little (Appendix Table~\ref{tab:step1-screen}).

\paragraph{Step-0 rank versus trajectory shift.} Block-3 \texttt{mlp.proj} has similarly low step-0 rank (0.36\%, Table~\ref{tab:rank}) but only $1.4\times$ rank gain, versus $7.8\times$ for the adapter. The contrast shows that low step-0 rank alone does not measure trajectory shift.

Ouro-1.4B has no Huginn-style adapter, yet its layer-0 \texttt{down\_proj} has rank 1/5{,}632 (0.02\%) at step~0, while per-step Hessians at steps 1--3 individually reach rank 974--1{,}194 (Appendix Table~\ref{tab:rank-survey}). Step~0 contributes 99.5\% of the 4-step trace sum, leaving the later directions below the accumulated Hessian's relative cutoff and its thresholded rank at 1. Their covariance still changes GPTQ's solution: 4-step calibration improves PPL from 14.67 to 12.97 (Table~\ref{tab:main}). Calibration blindness therefore extends beyond Huginn. Huginn's adapter is distinctive because it is both feedback-exposed and calibration-blind.

\paragraph{Method dependence.} AWQ~\citep{lin2024awq} selects per-channel scales from activation magnitudes, which are stable across recurrence steps (Pearson $r = 0.89$ between step~0 and step~31 on the Huginn adapter). AWQ is insensitive to calibration step (0.01 PPL range vs.\ 0.29 for GPTQ), but step-0 AWQ improves by only 0.06 PPL over RTN with the same asymmetric quantizer, comparable to AWQ's 0.01--0.18 PPL gains on standard LLMs at the same precision~\citep{lin2024awq} (Appendix Table~\ref{tab:awq}). Hessian-based solvers exploit the covariance subspace that per-channel magnitudes miss, but their reliance on $H$ makes them vulnerable to step-0 mismatch.

\subsection{Cross-Architecture Rank Survey}
\label{sec:rank-validation}

\begin{figure}[H]
\centering
\includegraphics[width=\linewidth]{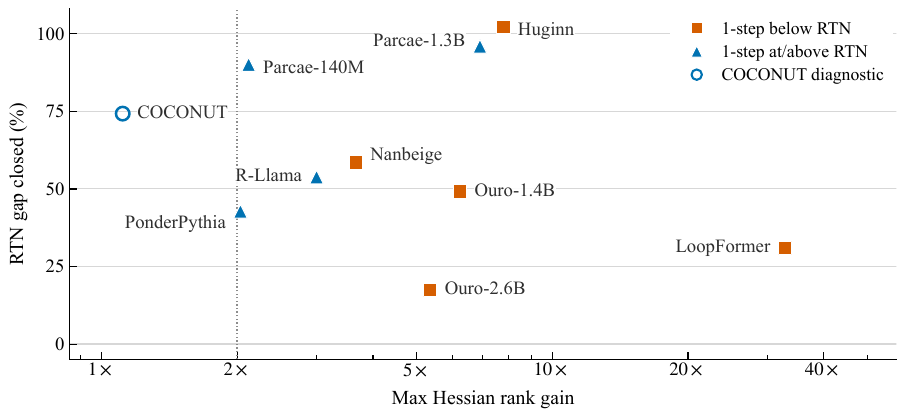}
\caption{Maximum accumulated-to-step-0 Hessian rank ratio across shared layers versus the RTN-to-full-precision gap closed by trajectory calibration. Marker shape shows whether one-step GPTQ outperforms RTN. The hollow COCONUT point is an in-sample top-1 diagnostic ($n{=}30$); all 9 filled points have rank gain $\geq 2\times$.}
\label{fig:rank-scatter}
\end{figure}

Rank gain is the accumulated-to-step-0 rank ratio for one shared layer, measuring how much subspace coverage grows after step~0. Figure~\ref{fig:rank-scatter} reports the maximum for each checkpoint and plots recovery separately on the vertical axis. LoopFormer and COCONUT start from comparable minimum step-0 rank ratios (1.5\% for LoopFormer \texttt{c\_fc} and 1.4\% for COCONUT \texttt{c\_proj}; Appendix Table~\ref{tab:rank-survey}), yet later steps add far more directions for LoopFormer. All nine checkpoints in Table~\ref{tab:main} have a maximum rank gain of at least $2\times$, so later steps add above-threshold directions across every architecture tested.

Rank gain is large when a layer's step-0 input differs qualitatively from its later-step input. Huginn's adapter concatenates a noise-initialized state with the prompt embedding. Ouro's layer-0 sees raw embeddings before any transformer block has shaped them. LoopFormer's iteration-conditioned adaLN~\citep{jeddi2026loopformer} changes layer inputs across loops. Among its GPTQ-quantized projections, the lowest-rank \texttt{c\_fc} gains $4.1\times$. By contrast, COCONUT reuses GPT-2's full residual stack across latent steps, and its per-step ranks barely shift ($43 \to 50$).

Model-level averages hide low-rank layers. Ouro-2.6B's worst layer (0.02\%) is masked by 47 near-full-rank layers, so an average would not flag it.

% Section 4 — Recurrence-Aware Calibration
% arXiv preprint

\section{Trajectory Calibration}
\label{sec:method}

Instead of calibrating at step 0 alone, we accumulate the Hessian across the full rollout. The step-0 Hessian assigns low relative weight to directions that later recurrence steps activate (Section~\ref{sec:calibration}). On the adapter, 99.6\% of input directions fall below the reported $10^{-3}\lambda_{\max}$ threshold, and GPTQ's column compensation can shift rounding error toward them. With the state trajectory fixed at the bf16 reference, the reconstruction proxy over the rollout is
\begin{equation}
  L_{\text{deploy}} \approx \sum_{t=0}^{N-1} \mathrm{tr}\bigl(\Delta W\, H_t\, \Delta W^\top\bigr) = \mathrm{tr}\bigl(\Delta W\, H_{\text{deploy}}\, \Delta W^\top\bigr),
  \label{eq:deploy}
\end{equation}
where $H_t = X_t^\top X_t$ and $H_{\text{deploy}} = \sum_t H_t$. One-step GPTQ computes column compensation from a damped step-0 Hessian: damping regularizes low-energy directions but cannot supply their later-step activation covariance. Accumulating across the rollout adds that covariance, so the solver penalizes rounding error along directions used during deployment.

\paragraph{Quantized-trajectory control.} At grouped INT3 ($g{=}128$), a stronger stress test than the main INT4 setting, recollecting Huginn's Hessians on the quantized rollout changes final-step top-1 agreement with bf16 from 84.59\% to 84.95\% on 20 prompts held out from calibration (prompt-cluster bootstrap 95\% CI, $[-2.31, 3.10]$\,pp). The grouped-INT4 reconstruction proxy is also nearly constant in magnitude along the bf16 rollout (maximum-to-minimum ratio $1.01\times$ for Huginn and $1.02\times$ for Ouro-1.4B), even as the active directions change (Appendix~\ref{sec:app-frozen}).

Consistent with the role of missing directions, COCONUT's thresholded Hessian rank changes little across steps (maximum gain $1.1\times$; Section~\ref{sec:rank-validation}), and trajectory accumulation improves its in-sample top-1 agreement by only 0.8 percentage points.

\begin{figure}[t]
\centering
\includegraphics[width=\linewidth]{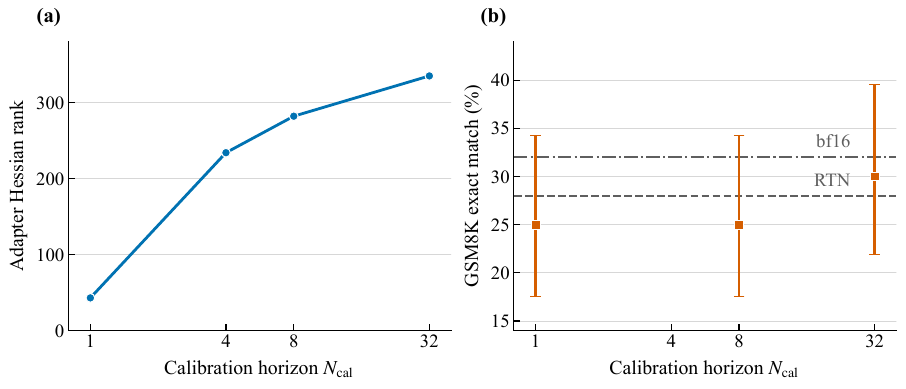}
\caption{Huginn-3.5B calibration horizon (grouped INT4, $g{=}128$). (a) Adapter Hessian rank. (b) Exact match on GSM8K test IDs 0--99, held out from calibration, with Wilson 95\% intervals and matched bf16/RTN references. Panels use separate runs.}
\label{fig:saturation}
\end{figure}

\paragraph{Single late-step calibration.} On Huginn WikiText-2, step-31-only calibration reaches 15.07 PPL under the same simulated grouped-INT4 setting, compared with 15.36 for one-step and 14.89 for full accumulation. It captures 62\% of the one-step-to-full improvement. Step~31 has higher thresholded rank (361 vs.\ 335 at $10^{-3}\lambda_{\max}$), so thresholded rank alone does not determine recovery.

\paragraph{Calibration horizon.} The gain appears only at the full 32-step horizon in the matched 100-example Huginn sweep. Exact-match accuracy is 25\% at $N_{\mathrm{cal}}{=}1$ and 8, then 30\% at $N_{\mathrm{cal}}{=}32$, compared with 28\% for RTN and 32\% for bf16 (Figure~\ref{fig:saturation}). Because horizon curves differ across architectures, as the Ouro and Parcae sweeps show (Appendix Table~\ref{tab:saturation-cross}), the main table uses each model's deployed recurrence depth.

% Section 5 — Experiments
% arXiv preprint

\section{Evaluation}
\label{sec:experiments}

Sections~\ref{sec:mechanism} and~\ref{sec:calibration} identify feedback exposure and calibration blindness under recurrent weight reuse. We test the first with controlled interventions and the second with rank and recovery results across nine checkpoints from Huginn, R-Llama, Ouro, LoopFormer, Parcae, PonderPythia~\citep{zeng2026ponderlm}, and Nanbeige~\citep{nanbeige2026nanbeige42}.

\paragraph{Setup.} All recovery rows use deterministic simulated grouped-INT4 ($g{=}128$). Huginn and R-Llama calibrate on the first 50 GSM8K training questions and report the disjoint full test set ($n{=}1{,}319$, 8-shot, greedy). The other checkpoints use WikiText-2 perplexity. Appendix Tables~\ref{tab:bpb} and~\ref{tab:supporting-metrics} give separate bits-per-byte reruns and supporting GSM8K/LAMBADA results. Table~\ref{tab:main} gives each GPTQ horizon; full details are in Appendix~\ref{sec:app-setup}.

\subsection{Performance Recovery}
\label{sec:main-results}

If the failure comes from directions underweighted by the step-0 Hessian, accumulating
Hessians along the recurrence should recover performance. Table~\ref{tab:main} tests this prediction across nine checkpoints.

\begin{table}[!htbp]
\centering
\caption{Simulated grouped-INT4 recovery ($g{=}128$). $N_{\mathrm{cal}}$ is the calibration horizon; gap closed is $(N\text{-step}-RTN)/(Base-RTN)$, computed from unrounded results. Base is bf16 except Parcae (f32). The best quantized result in each row is bold.}
\label{tab:main}
\begingroup
\small
\setlength{\tabcolsep}{2.6pt}
\renewcommand{\arraystretch}{1.04}
\begin{tabular}{@{}lcrrrrr@{}}
\toprule
\textbf{Checkpoint} & $\boldsymbol{N_{\mathrm{cal}}}$ & \textbf{Base} & \textbf{RTN} & \textbf{1-step} & \textbf{$N$-step} & \textbf{Gap closed} \\
\midrule
\multicolumn{7}{@{}l}{\emph{Full GSM8K exact-match accuracy $\uparrow$}} \\
Huginn 3.5B\textsuperscript{a} & 32 & 34.9 & 27.9 & 26.9 & \textbf{35.0} & 102\% \\
R-Llama 1.4B & 32 & 50.2 & 39.5 & 40.3 & \textbf{45.3} & 54\% \\
\midrule
\multicolumn{7}{@{}l}{\emph{Full WikiText-2 perplexity $\downarrow$}} \\
Ouro 1.4B & 4 & 11.51 & 14.38 & 14.67 & \textbf{12.97} & 49\% \\
Ouro 2.6B\textsuperscript{b} & 4 & 10.10 & 11.94 & 13.61 & \textbf{11.62} & 17\% \\
LoopFormer 278M & 8 & 31.80 & 35.10 & 45.91 & \textbf{34.08} & 31\% \\
Parcae 140M & 8 & 38.46 & 46.08 & 41.49 & \textbf{39.21} & 90\% \\
Parcae 1.3B & 8 & 19.57 & 51.30 & 27.35 & \textbf{20.84} & 96\% \\
PonderPythia 2.8B & 4 & 11.75 & 12.61 & 12.51 & \textbf{12.24} & 43\% \\
Nanbeige4.2-3B-Base\textsuperscript{c} & 2 & 10.50 & 13.13 & 13.23 & \textbf{11.59} & 58\% \\
\bottomrule
\end{tabular}
\par\smallskip
{\footnotesize\raggedright\textsuperscript{a} No rotation; exact counts are 460/1{,}319 for bf16, 368/1{,}319 for RTN, and 462/1{,}319 for $N$-step, giving $94/92=102\%$. \textsuperscript{b} The 17\% closure corresponds to 1.99 PPL over one-step GPTQ and 0.32 over RTN. \textsuperscript{c} Unrounded Nanbeige PPL is 10.5033/13.1284/11.5932 for bf16/RTN/$N$-step, giving 58.48\%; the model retains its native 131{,}072-token context.\par}
\endgroup
\end{table}

One-step GPTQ underperforms RTN on five checkpoints; on LoopFormer, the deficit is 10.81 PPL. Trajectory calibration improves every row: Huginn returns to bf16-level accuracy, and the others close 17--96\% of the RTN-to-base gap. On Huginn, $N$-step calibration with rotation also remains within 2\,pp of bf16 on ARC-C and HellaSwag (Appendix Table~\ref{tab:benchmarks}).

\paragraph{Architecture and scale.}

Parcae has no adapter, providing a test of calibration blindness without Huginn's loop-entry adapter. At 140M, trajectory calibration closes 90\% of the RTN-to-base gap. At 1.3B, it lowers PPL from 27.35 with one-step GPTQ to 20.84, compared with 51.30 for RTN and 19.57 for f32. One projection falls to 0.03\% step-0 rank despite the model's ${\sim}$16\% core-block average (Appendix Table~\ref{tab:rank-survey}). LAMBADA accuracy confirms the ranking (Appendix Table~\ref{tab:supporting-metrics}).

On Ouro-2.6B, trajectory calibration improves PPL by 1.99 over one-step GPTQ and 0.32 over RTN. RTN is already within 1.84 PPL of base. GSM8K confirms the ordering: 79.8\% for 4-step calibration, 78.8\% for RTN, and 72.6\% for one-step (Appendix Table~\ref{tab:supporting-metrics}).

\begin{samepage}
On PonderPythia-2.8B, four-pass calibration lowers PPL from 12.51 to 12.24 (RTN: 12.61). Nanbeige4.2-3B extends the result to 4.17B total parameters: one-step GPTQ is worse than RTN, while 2-pass calibration closes 58\% of the RTN-to-bf16 gap at its deployed depth of two.
\end{samepage}

\paragraph{Additional bit widths.}

On WikiText-2, per-channel INT8 leaves Huginn within 0.01 PPL of bf16 (Appendix~\ref{sec:app-rotation-type}). On the full GSM8K test set at grouped INT3, exact match rises from 5.84\% with one-step GPTQ to 20.70\% with full-trajectory calibration. RTN reaches 8.11\%, while equal-row trajectory calibration reaches 17.44\% with the same activation-row budget as one-step (Appendix Table~\ref{tab:int3}).

\subsection{Real-Kernel Evaluation}
\label{sec:system}

Trajectory calibration preserves the inference graph. On a single NVIDIA A100 at batch size 1, median eager latency for a 32-step Huginn forward on a 7-token prompt is 303.9\,ms in bf16 and 250.9\,ms with the Marlin W4A16 kernel ($g{=}128$), a $1.21\times$ speedup (paired-bootstrap 95\% CI, $1.20$--$1.23\times$). Across inputs of 1, 7, and 16 tokens, the speedup ranges from $1.21\times$ to $1.23\times$. The 17 shared projections use 3.27\,GB for bf16 weights and 0.86\,GB in Marlin format, a $3.80\times$ reduction. With the same quantized weights, exact-match accuracy on the full GSM8K test set is 34.0\% under Marlin and 35.0\% under deterministic simulated INT4 (Appendix~\ref{sec:app-rotation-type}).

% Section 6 — Related Work
% arXiv preprint

\section{Related Work}
\label{sec:related}

Hyperloop Transformers studies loop-level hyper-connections. In one weight-only INT4 experiment, it accumulates GPTQ Hessians across loops without a step-0 comparison or subspace analysis~\citep{zeitoun2026hyperlooptransformers}. LoopQ attributes failure to loop-dependent distributions, state reuse, and recursive error; it learns loop-dependent scales and transforms, a transition adapter, and logit/hidden-state distillation under W4A4/W4A8~\citep{fang2026loopqquantizationrecursivetransformers}. \citet{jim2026recursivecompression} report architecture-dependent INT4 fragility in small recursive reasoners. Our causal analysis separates the mechanisms. Rank and subspace measurements identify directions missed by step-0 Hessians, while controlled changes to prior-error feedback and the identity path isolate error recirculation and residual protection.

\paragraph{Calibration mismatch in iterative systems.}
Q-Diffusion samples denoising timesteps, while AdaTSQ uses timestep-dependent activation policies and temporally weighted calibration~\citep{qdiffusion2023,adatsq2024}. In a looped transformer, the same operator generates and processes later states, so a step-0 Hessian can miss directions used during deployment (Section~\ref{sec:calibration}).

\paragraph{Quantized recurrence.}
Quamba's contractive-LTI bound assumes an unperturbed $A$~\citep{quamba2024}. Our IIR and Mamba controls perturb the transition itself; in the IIR control, quantizing $B$, $C$, and $D$ leaves the pole spectrum fixed (Section~\ref{sec:mechanism}). QMamba ablates SSM components, while Bi-Mamba keeps SSM parameters at full precision and binarizes input/output projections~\citep{qmamba2024,bimamba2024}. Monotone DEQ theory bounds operator perturbations under a certified monotonicity margin~\citep{li2026mondeq}. Precision highways address quantization during training, at the cost of retraining~\citep{park2018highway}.

\paragraph{Residual connections and error propagation.}
Residual-network stability and path-ensemble analyses do not cover repeated re-entry of the same perturbation~\citep{haber2017stable,veit2016residual}. Our Jacobian analysis extends the stability view to discrete recurrence. The identity-bypass control isolates residual protection under fixed evaluation states and adapter weights (Section~\ref{sec:residual}).

\paragraph{PTQ methods.}
GPTQ, AWQ, SpinQuant, QuIP\#, and SmoothQuant~\citep{frantar2022gptq,lin2024awq,ashkboos2024spinquant,chee2024quip,xiao2023smoothquant} typically use single-pass rather than recurrence-conditioned calibration. With step-0 calibration, GPTQ and LDLQ~\citep{chee2023quip} are equivalent adaptive-rounding algorithms that minimize $\mathrm{tr}(\Delta W\, H\, \Delta W^\top)$, so both inherit $H$'s rank deficiency. QuIP\#'s Hadamard rotation is orthogonal and rank-preserving, so it does not restore missing directions in $H$ (Appendix Table~\ref{tab:ablation}). AWQ selects per-channel scales from activation magnitudes, which are more stable across recurrence steps (Section~\ref{sec:calibration}, Appendix Table~\ref{tab:awq}).

% Section 7 — Conclusion
% arXiv preprint

\section{Conclusion}
\label{sec:conclusion}

Conventional calibration-based PTQ uses activations from a single forward pass. In looped transformers, weight reuse turns quantization into a feedback problem: rounding error becomes dangerous when it enters the recurrent state transition without an identity path. IIR and Mamba controls likewise show greater damage when quantization changes the recurrent transition.

Calibration faces a parallel problem. A step-0 Hessian can miss the directions used later in the rollout, and one-step GPTQ underperforms RTN on five of nine checkpoints. Accumulating Hessians along the recurrence trajectory recovers performance in every tested case.

\noindent\textbf{Limitations.} Our evaluation covers nine checkpoints up to 4.17B parameters. Trajectory calibration does not always recover full-precision performance, and rank gain alone does not predict recovery magnitude.

We expect the same failure modes in larger looped models that retain unprotected feedback paths or step-dependent activation subspaces. For looped models, a quantizer that never sees the loop should not be trusted to quantize it.

\bibliographystyle{assets/plainnat}
\bibliography{references}

\clearpage
\beginappendix
% Appendix
% arXiv preprint

\section{Architecture Diagram}
\label{sec:app-architecture}

\begin{figure}[!htbp]
\centering
\begin{tikzpicture}[
  >=stealth, thick,
  block/.style={draw, rounded corners=2pt, minimum width=1.8cm, minimum height=0.7cm, align=center, font=\small},
  arr/.style={->, thick},
]
% Adapter (no residual)
\node[block, fill=red!12] (adp) at (0,0) {Adapter\\[-1pt]{\scriptsize(no residual)}};
% Core blocks (with residual)
\node[block, fill=green!12] (c1) at (3.0,0) {Core $1$\\[-1pt]{\scriptsize$f(x)+x$}};
\node[block, fill=green!12] (c4) at (6.0,0) {Core $4$\\[-1pt]{\scriptsize$f(x)+x$}};
\node at (4.5,0) {$\cdots$};
% Input embeddings
\node[font=\small] (ie) at (0,1.4) {$e$ (input emb.)};
% State labels
\node[font=\small] (st) at (-2.2,0) {$s_t$};
\node[font=\small] (st1) at (8.2,0) {$s_{t+1}$};
% Arrows
\draw[arr] (st) -- (adp);
\draw[arr] (ie) -- (adp);
\draw[arr] (adp) -- (c1);
\draw[arr] (c1) -- ++(0.7,0);
\draw[arr] (5.0,0) -- (c4);
\draw[arr] (c4) -- (st1);
% Feedback loop
\draw[arr, dashed] (st1.south) -- ++(0,-0.7) -| (st.south);
% cat label
\node[font=\scriptsize, anchor=west] at (-0.15,0.7) {cat};
\end{tikzpicture}
\caption{Huginn recurrence loop. The adapter projects $\mathrm{cat}(s_t, e)$ back to the model width with no residual path. Adapter perturbations therefore enter the recurrence without an identity bypass, while core residual paths provide an identity channel that can partially buffer small perturbations.}
\label{fig:architecture}
\end{figure}
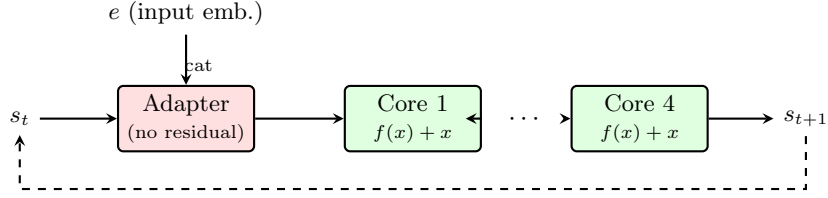

\section{Trajectory Calibration Algorithm}

\begin{algorithm}[!htbp]
\caption{Trajectory calibration for looped transformers.}
\label{alg:trajectory}
\begin{algorithmic}[1]
\REQUIRE bf16 model, prompts $\mathcal{D}$, recurrence depth $N$
\FOR{each shared layer $\ell$}
  \STATE $H_\ell \gets \sum_{t=0}^{N-1} {(X_t^\ell)}^\top X_t^\ell$
  \STATE $\hat{W}_\ell \gets \text{GPTQ}(W_\ell, H_\ell)$ \hfill $\triangleright$ unchanged column solver
\ENDFOR
\end{algorithmic}
\end{algorithm}

\section{Supporting Tables}
Tables~\ref{tab:controlled}--\ref{tab:ablation} collect supplementary numeric
results referenced in Sections~\ref{sec:mechanism} and~\ref{sec:experiments}.

\begin{table}[!htbp]
\centering
\caption{Controlled injection for Huginn-3.5B (50 prompts; data from Figure~\ref{fig:hero}). Arms 1--3 use deterministic per-channel INT4 FakeQuant on the listed layers, with all other layers bf16. Arm 4 adds one fixed Gaussian weight-noise draw with the same Frobenius norm as the adapter's per-channel INT4 weight error. Two bf16 runs with independently sampled initial states reach 98.9\% agreement.}
\label{tab:controlled}
\small
\begin{tabular}{@{}llrl@{}}
\toprule
\textbf{Arm} & \textbf{Quantized layers} & \textbf{Top-1 (N{=}32)} & \textbf{Top-1 endpoints ($N{=}1 \rightarrow 32$)} \\
\midrule
0 & None (bf16 control) & 98.9\% & 68.8\% $\to$ 98.9\% \\
\midrule
1 & Adapter only & \textbf{10.5\%} & 24.8\% $\to$ 10.5\% \\
2 & Core only (16 projections) & \textbf{87.9\%} & 68.4\% $\to$ 87.9\% \\
3 & All (adapter + core) & 10.6\% & 25.0\% $\to$ 10.6\% \\
\midrule
4 & Gaussian noise (norm-matched) & 25.0\% & 49.3\% $\to$ 25.0\% \\
\bottomrule
\end{tabular}
\end{table}

\begin{table}[!htbp]
\centering
\caption{Controlled injection for R-Llama-1.4B (50 prompts, per-channel INT4 FakeQuant). The adapter/core split replicates on a second architecture: adapter-only INT4 drops agreement to 49.1\%, while core-only preserves 74.8\%. The gap is narrower than Huginn (25.7\,pp vs.\ 77.4\,pp). Two bf16 runs with independently sampled initial states reach 98.7\% agreement.}
\label{tab:controlled-rllama}
\small
\begin{tabular}{@{}llr@{}}
\toprule
\textbf{Arm} & \textbf{Quantized layers} & \textbf{Top-1 (N{=}32)} \\
\midrule
0 & None (bf16 control) & 98.7\% \\
\midrule
1 & Adapter only & \textbf{49.1\%} \\
2 & Core blocks only (6 layers) & \textbf{74.8\%} \\
3 & All (adapter + core) & 34.6\% \\
\bottomrule
\end{tabular}
\end{table}

\begin{table}[!htbp]
\centering
\caption{Huginn-3.5B per-channel INT4 diagnostic. Top-1 agreement uses 20 prompts disjoint from calibration. GSM8K uses the first 100 test examples. For trajectory GPTQ, those 100 GSM8K inputs are also used without labels for Hessian calibration, so the corresponding task score is diagnostic.}
\label{tab:perchannel-diagnostic}
\small
\begin{tabular}{@{}lcc@{}}
\toprule
\textbf{Configuration} & \textbf{Top-1 (N{=}32)} & \textbf{GSM8K strict/flex} \\
\midrule
bf16 & $\sim$100\% & 34\% / 41\% \\
Per-channel INT4, no rotation & 8.5\% & 0\% / 0\% \\
Per-channel RTN + rotation & 82.6\% & 13\% / 15\% \\
Per-channel $N$-step + rotation & \textbf{92.7\%} & \textbf{24\% / 25\%} \\
\bottomrule
\end{tabular}
\end{table}

\begin{table}[!htbp]
\centering
\caption{Huginn-3.5B calibration-horizon diagnostics (deterministic grouped-INT4 FakeQuant, $g{=}128$, no rotation). The upper block reports Hessian rank and top-1 agreement. The lower block is a separate held-out GSM8K evaluation on test examples 0--99, held fixed across rows. For comparison, the separate step-31-only Hessian has rank 361 (3.42\%), while the 32-step accumulated Hessian has rank 335. Each is evaluated against its own relative threshold (Section~\ref{sec:method}).}
\label{tab:saturation}
\small
\begin{tabular}{@{}rrrc@{}}
\toprule
\multicolumn{4}{@{}l}{\emph{Rank and top-1 diagnostic}} \\
\textbf{$N_{\mathrm{cal}}$} & \textbf{Hessian rank} & \textbf{Rank (\% of dim.)} & \textbf{Top-1 (32-step)} \\
\midrule
1  & 43  & 0.41\% & 88.2\% \\
4  & 234 & 2.22\% & 90.9\% \\
8  & 282 & 2.67\% & \textbf{92.5\%} \\
32 & \textbf{335} & \textbf{3.17\%} & \textbf{92.5\%} \\
\midrule
\multicolumn{3}{@{}l}{RTN baseline} & 88.0\% \\
\bottomrule
\end{tabular}
\par\smallskip
\begin{tabular}{@{}lrr@{}}
\toprule
\multicolumn{3}{@{}l}{\emph{Clean held-out GSM8K, same 100 test examples}} \\
\textbf{Configuration} & \textbf{Strict} & \textbf{Flexible} \\
\midrule
bf16 & 32/100 & 38/100 \\
RTN & 28/100 & 32/100 \\
$N_{\mathrm{cal}}{=}1$ & 25/100 & 28/100 \\
$N_{\mathrm{cal}}{=}8$ & 25/100 & \textbf{33/100} \\
$N_{\mathrm{cal}}{=}32$ & \textbf{30/100} & 31/100 \\
\bottomrule
\end{tabular}
\end{table}

\begin{table}[!htbp]
\centering
\caption{Calibration horizons across architectures (grouped-INT4 FakeQuant, $g{=}128$). Each row reports one calibration horizon. Ouro-1.4B obtains 88\% of the $N{=}1$-to-$N{=}4$ PPL improvement by $N_{\mathrm{cal}}{=}2$. Parcae-1.3B improves sharply by $N_{\mathrm{cal}}{=}4$; at $N_{\mathrm{cal}}{=}2$, one layer has only 3 eigenvalues above the rank threshold and PPL is worse than RTN.}
\label{tab:saturation-cross}
\small
\begin{tabular}{@{}llrr@{}}
\toprule
\textbf{Model} & $\boldsymbol{N_{\mathrm{cal}}}$ & \textbf{WikiText-2 PPL} & \textbf{LAMBADA acc} \\
\midrule
Ouro 1.4B & 1 & 14.67 & --- \\
& 2 & 13.17 & --- \\
& 3 & 13.05 & --- \\
& 4 & \textbf{12.97} & --- \\
\midrule
Parcae 1.3B & 1 & 27.35 & 0.341 \\
& 2 & 99.59$^{\dagger}$ & 0.326 \\
& 4 & 20.98 & \textbf{0.430} \\
& 6 & \textbf{20.62} & --- \\
& 8 & 20.84 & 0.416 \\
\bottomrule
\end{tabular}
\par\smallskip
{\footnotesize\raggedright $^{\dagger}$ At $N_{\mathrm{cal}}{=}2$, block-0 \texttt{mlp.proj} has only 3/6{,}144 eigenvalues above $10^{-3}\lambda_{\max}$, and GPTQ gives worse PPL than RTN (99.59 versus 51.30). At $N_{\mathrm{cal}}{=}4$, PPL recovers to 20.98. PPL is not monotone thereafter, increasing slightly from 20.62 at $N_{\mathrm{cal}}{=}6$ to 20.84 at $N_{\mathrm{cal}}{=}8$.\par}
\end{table}

\begin{table}[!htbp]
\centering
\caption{Auxiliary Huginn/R-Llama benchmarks for $g{=}128$ with trajectory calibration plus adapter rotation. ARC-C and HellaSwag are accuracy (higher is better); WikiText-2 is perplexity (lower is better). The main-table rows (Table~\ref{tab:main}) isolate trajectory calibration without rotation. For comparison, Section~\ref{sec:method} reports 14.89 PPL for the non-rotated Huginn trajectory-calibration run.}
\label{tab:benchmarks}
\small
\begin{tabular}{@{}llrrr@{}}
\toprule
\textbf{Model} & \textbf{Config} & \textbf{ARC-C (\%)} & \textbf{HellaSwag (\%)} & \textbf{WikiText-2 PPL} \\
\midrule
Huginn 3.5B & bf16 & 37.4\% & 66.6\% & 14.02 \\
& $N$-step + rotation & 37.0\% & 65.3\% & 14.54 \\
\midrule
R-Llama 1.4B & bf16 & 37.1\% & 46.1\% & 30.49 \\
& $N$-step + rotation & 36.3\% & 45.2\% & 32.64 \\
\bottomrule
\end{tabular}
\end{table}

\begin{table}[!htbp]
\centering
\caption{Supporting metrics for Ouro-2.6B and Parcae-1.3B. Table~\ref{tab:main} uses WikiText-2 perplexity as the primary metric for these checkpoints, and the GSM8K and LAMBADA results preserve the same method ordering.}
\label{tab:supporting-metrics}
\small
\begin{tabular}{@{}llccc@{}}
\toprule
\textbf{Checkpoint} & \textbf{Metric} & \textbf{RTN} & \textbf{1-step} & \textbf{$N$-step} \\
\midrule
Ouro-2.6B & GSM8K strict/flex ($n{=}1{,}319$) & 78.8 / 81.5 & 72.6 / 74.6 & \textbf{79.8 / 82.8} \\
Parcae-1.3B & LAMBADA accuracy & 0.3117 & 0.3410 & \textbf{0.4157} \\
\bottomrule
\end{tabular}
\par\smallskip
{\footnotesize Parcae-1.3B f32 LAMBADA accuracy is 0.4584.}
\end{table}

\begin{table}[!htbp]
\centering
\caption{Full-test Huginn GSM8K at grouped INT3 ($g{=}128$; $n{=}1{,}319$). Equal-row calibration uses the one-step activation-row budget distributed across all 32 recurrence steps.}
\label{tab:int3}
\small
\begin{tabular}{@{}lrr@{}}
\toprule
\textbf{Configuration} & \textbf{Strict (\%)} & \textbf{Flexible (\%)} \\
\midrule
bf16 & 34.87 (460/1{,}319) & 42.23 (557/1{,}319) \\
RTN & 8.11 (107/1{,}319) & 8.64 (114/1{,}319) \\
1-step GPTQ & 5.84 (77/1{,}319) & 6.44 (85/1{,}319) \\
Equal-row trajectory GPTQ & 17.44 (230/1{,}319) & 17.82 (235/1{,}319) \\
Full-trajectory GPTQ & \textbf{20.70} (273/1{,}319) & \textbf{23.28} (307/1{,}319) \\
\bottomrule
\end{tabular}
\par\smallskip
{\footnotesize\raggedright Full trajectory improves over one-step by 14.9\,pp strict and 16.8\,pp flexible (paired-bootstrap 95\% CIs [12.66, 17.06] and [14.56, 19.18]; exact McNemar $p=9.35\times10^{-39}$ and $2.72\times10^{-45}$, respectively). Against equal-row calibration, the gains are 3.3\,pp strict and 5.5\,pp flexible; the strict paired-bootstrap 95\% CI is [0.83, 5.69], with exact McNemar $p=0.00975$.\par}
\end{table}

\begin{table}[!htbp]
\centering
\caption{Direct bits per byte on WikiText-2 for the seven perplexity checkpoints in Table~\ref{tab:main}; lower is better. Base is bf16 except for Parcae (f32). The best quantized result in each row is bold. All Nanbeige conditions retain the checkpoint's native 131{,}072-token context configuration.}
\label{tab:bpb}
\small
\setlength{\tabcolsep}{3pt}
\begin{tabular}{@{}lrrrr@{}}
\toprule
\textbf{Checkpoint} & \textbf{Base} & \textbf{RTN} & \textbf{1-step} & \textbf{$N$-step} \\
\midrule
Ouro 1.4B & 0.6591 & 0.7193 & 0.7238 & \textbf{0.6912} \\
Ouro 2.6B & 0.6240 & 0.6689 & 0.7014 & \textbf{0.6582} \\
LoopFormer 278M & 0.9334 & 0.9600 & 1.0327 & \textbf{0.9516} \\
Parcae 140M & 0.9847 & 1.0326 & 1.0053 & \textbf{0.9899} \\
Parcae 1.3B & 0.8032 & 1.0514 & 0.8895 & \textbf{0.8311} \\
PonderPythia 2.8B & 0.6646 & 0.6837 & 0.6817 & \textbf{0.6757} \\
Nanbeige4.2-3B-Base & 0.6345 & 0.6947 & 0.6968 & \textbf{0.6611} \\
\bottomrule
\end{tabular}
\end{table}

\begin{table}[H]
\centering
\caption{Component ablation on Huginn-3.5B ($g{=}128$, $n{=}100$). Scores are diagnostic because GPTQ uses these unlabeled inputs for Hessian calibration. Bold marks the feedback-protection and best-quantized results.}
\label{tab:ablation}
\small
\begin{tabular}{@{}lrr@{}}
\toprule
\textbf{Configuration} & \textbf{Strict (\%)} & \textbf{Flex (\%)} \\
\midrule
RTN (no rotation, no GPTQ) & 25\% & 28\% \\
1-step GPTQ (rotated variant) & 25\% & 28\% \\
\textbf{Adapter-bf16 + RTN (no rotation)} & \textbf{32\%} & \textbf{35\%} \\
\midrule
Adapter $N$-step + rotation (core 1-step) & 23\% & 29\% \\
$N$-step all layers (no rotation) & 30\% & 37\% \\
$N$-step all layers (rotated variant) & \textbf{34\%} & \textbf{39\%} \\
\midrule
bf16 & 34\% & 41\% \\
\bottomrule
\end{tabular}
\par\smallskip
{\footnotesize\raggedright Rotated rows bundle adapter rotation with the calibration variant used in that ablation. ``Adapter $N$-step + rotation'' applies trajectory calibration and rotation to the adapter while the core uses one-step calibration.\par}
\end{table}

\begin{table}[H]
\centering
\caption{Calibration-step sensitivity on Huginn-3.5B WikiText-2 ($g{=}128$, deterministic FakeQuant). Bold marks the best result within each method family.}
\label{tab:awq}
\small
\begin{tabular}{@{}lc@{}}
\toprule
\textbf{Configuration} & \textbf{WikiText-2 PPL} \\
\midrule
RTN $g{=}128$ & 14.98 \\
\midrule
\multicolumn{2}{@{}l}{\emph{GPTQ}} \\
\quad Step 0 (1-step) & 15.36 \\
\quad Step 31 & 15.07 \\
\quad 32-step & \textbf{14.89} \\
\midrule
\multicolumn{2}{@{}l}{\emph{AWQ}} \\
\quad Step 0 & 14.66 \\
\quad Step 31 & \textbf{14.65} \\
\quad All steps & \textbf{14.65} \\
\bottomrule
\end{tabular}
\par\smallskip
{\footnotesize\raggedright AWQ uses per-channel activation magnitudes, while GPTQ uses $H=X^\top X$. GPTQ and RTN are symmetric. AWQ is asymmetric (asymmetric RTN: 14.72 PPL), so absolute values are not directly comparable across method families.\par}
\end{table}

\section{Rotation and Marlin Implementation Checks}
\label{sec:app-rotation-type}

All rotation comparisons use adapter-only rotation (\texttt{--no-core-rotation}). The simulated evaluations are deterministic, while the Marlin deployment diagnostic below uses the real kernel.
Table~\ref{tab:rotation-perchannel} reports the per-channel diagnostic where rotation matters most.

\begin{table}[!htbp]
\centering
\caption{Hadamard vs.\ random orthogonal at per-channel INT4 with $N$-step GPTQ (both at block size 512). Top-1 agreement uses 20 prompts disjoint from calibration. The GSM8K task evaluations include the unlabeled inputs used for Hessian calibration and are reported as diagnostics.}
\label{tab:rotation-perchannel}
\small
\begin{tabular}{@{}lccc@{}}
\toprule
\textbf{Config} & \textbf{Top-1 (N{=}32)} & \textbf{\shortstack{GSM8K \\ ($n{=}100$) strict/flex}} & \textbf{\shortstack{GSM8K \\ ($n{=}1{,}319$) strict/flex}} \\
\midrule
Random orthogonal & \textbf{91.0\%} & 23\% / 24\% & 26.9\% / 27.5\% \\
Hadamard & 90.7\% & \textbf{29\% / 33\%} & \textbf{29.4\% / 34.0\%} \\
bf16 & $\sim$100\% & 34\% / 41\% & 34.9\% / 42.2\% \\
\bottomrule
\end{tabular}
\end{table}

At INT8 per-channel RTN, Hadamard (block size 512) and random orthogonal both match bf16-level perplexity (14.02), while Hadamard with block size 32 gives 14.03. The QR fallback for partial Hadamard blocks removes the remaining INT8 perplexity difference; all tested INT8 rotation variants are within 0.01 PPL of bf16.

At grouped deployment precision, Marlin and deterministic FakeQuant use identical quantized weights and scales across all 17 shared projections. The no-rotation runs calibrate on 50 GSM8K training questions and evaluate all 1{,}319 test questions. Marlin reaches 34.0\%/34.3\% strict/flexible accuracy, versus 35.0\%/35.3\% for FakeQuant; the paired differences are $-1.0$/$-1.1$ percentage points (exact McNemar $p=0.28/0.24$).

The latency benchmark uses 10 warmups per arm and 50 paired bf16/Marlin trials in alternating order at each input length.

We also compare every shared Marlin projection against its dequantized grouped-INT4 reference at 1, 7, and 16 input tokens. Across these 51 comparisons, the maximum relative $\ell_2$ error is $1.35\times10^{-3}$. A 32-step full forward at each input length executes the expected 544 Marlin calls.

\section{Extended Controlled Injection Analysis}
\label{sec:app-arms}

\paragraph{Varying feedback strength.}
For each bf16 state $s_t^{\mathrm{bf16}}$ and current intervention state $s_t^\lambda$, we feed the quantized recurrence $s_t^{\mathrm{bf16}}+\lambda(s_t^\lambda-s_t^{\mathrm{bf16}})$. Thus $\lambda=0$ removes prior state error before each update, while $\lambda=1$ reproduces the natural quantized rollout. We use GSM8K training prompts 70--119 and 3 recurrent-state initializations shared across $\lambda\in\{0,0.25,0.5,0.75,1\}$.

We measure final relative state error as $\|s_{32}^\lambda-s_{32}^{\mathrm{bf16}}\|_2/\|s_{32}^{\mathrm{bf16}}\|_2$. With the full adapter quantized, the geometric mean of its prompt-level $\lambda=1$-to-$\lambda=0$ ratio is $1.989\times$ under per-channel INT4 (prompt-bootstrap 95\% CI, $[1.980,1.998]$) and $2.085\times$ at $g{=}128$ ($[1.971,2.229]$). Under per-channel INT4, cutting feedback raises final-step top-1 agreement with bf16 from 10.4\% to 69.2\%. Mean final state error increases monotonically with $\lambda$ in both settings, as does every prompt-level curve.

For a prespecified equal-width split, we quantize the full adapter once, retain its quantized values in either the 5{,}280 recurrent-state columns or 5{,}280 embedding columns, and leave the other half in bf16. Both arms show feedback amplification, but the embedding-half ratio is larger: $2.010\times$ versus $1.358\times$ under per-channel INT4, and $2.264\times$ versus $2.065\times$ at $g{=}128$. Both errors enter the recurrence through the same adapter output, so this split does not compare feedback-exposed with protected paths.

\paragraph{Adapter contribution under full INT4.}
Arm~1 (adapter INT4, core bf16) and Arm~3 (both INT4) produce nearly identical top-1 agreement (10.5\% vs.\ 10.6\%). Core INT4 noise adds little on top of adapter error. The adapter term remains dominant whether or not the core blocks are quantized, consistent with the controlled injection result in the main text (Table~\ref{tab:controlled}).

\section{Perturbation-Type Generalization}
\label{sec:app-perturbation-types}

The feedback-exposure mechanism is not specific to quantization noise. Table~\ref{tab:perturbation-types} reports separate weight-perturbation controls at $N{=}32$. The rank-4 rows use a common target norm ($\|\Delta W\|_F = 8.59$) at both locations to compare adapter and core perturbations at fixed size; their core arms split this norm between block-0 \texttt{mlp.fc} and \texttt{mlp.proj}. The trained directions use 50 next-token cross-entropy updates on the GSM8K training split before rescaling. For the Gaussian row, 5 random-input measurements of grouped-INT4 output error give target norms of 3.16 for the adapter and 8.11 for block-0 \texttt{mlp.fc}. Each perturbed forward is paired with its bf16 reference using the same initial-state seed.

\begin{table}[!htbp]
\centering
\caption{Perturbation-type generalization under the perturbation scales defined above. Top-1 agreement with bf16 at $N{=}32$. The trained rank-4 update is evaluated on 60 prompts; the random rank-4 update on 60 prompts for each of 3 seeds; Gaussian noise on 10 prompts for each of 3 seeds.}
\label{tab:perturbation-types}
\small
\begin{tabular}{@{}lccc@{}}
\toprule
\textbf{Perturbation} & \textbf{Adapter} & \textbf{Core} & \textbf{Gap (Core $-$ Adapter)} \\
\midrule
Trained rank-4 update & 13.4\% & 77.8\% & \textbf{64.4\,pp} \\
Random rank-4 update  & 71.8\% & 95.8\% & \textbf{24.0\,pp} \\
Gaussian noise     & 89.3\% & 96.5\% & \textbf{7.2\,pp} \\
\bottomrule
\end{tabular}
\par\smallskip
{\footnotesize Each Gaussian seed samples one fixed weight perturbation and reuses it for all 10 prompts. The bf16 and perturbed forwards use the same initial state.}
\end{table}

The random rank-4 update preserves a 24.0\,pp adapter/core gap (71.8\% vs.\ 95.8\%), and the trained update widens the gap to 64.4\,pp (13.4\% vs.\ 77.8\%). The Gaussian control gives the same ordering even though its adapter perturbation has the smaller norm (3.16 vs.\ 8.11). Core-block perturbations are less damaging in all three settings, consistent with buffering from the residual path.

\section{Sensitivity Metric Comparison}
\label{sec:app-sensitivity}
Table~\ref{tab:sensitivity-ranking} compares three per-projection sensitivity
measures. \emph{Ablation sensitivity} quantizes a single projection to grouped INT4
($g{=}128$) while all other projections remain at bf16, then reports the fraction of per-token
top-1 predictions that disagree with the all-bf16 baseline, averaged across
all 32 recurrence steps.
Higher values indicate that the projection's quantization error changes more
predictions. Random-direction gain is the arithmetic mean of
$\|f(x+\delta)-f(x)\|_F/\|\delta\|_F$ over 10 prompts and 10 Gaussian
perturbations per prompt, using the last captured layer input from a 16-step
rollout and $\delta=10^{-3}\|x\|_F z$ for $z\sim\mathcal{N}(0,I)$.
The third measure is $\mathrm{tr}(H)$, where $H=X^\top X$ is the activation Hessian used by GPTQ.

\begin{table}[!htbp]
\centering
\caption{Top five projections by ablation sensitivity from rankings across 17 shared projections under single-projection grouped INT4 ablation ($g{=}128$; 10 prompts, averaged across 32 steps). Ablation sensitivity is the fraction of per-token top-1 predictions that disagree with bf16 (0 = no effect, 1 = all predictions change). The adapter is a $2.3\times$ outlier by ablation but ranks last by random-direction gain and third by Hessian trace.}
\label{tab:sensitivity-ranking}
\small
\begin{tabular}{@{}lccc@{}}
\toprule
\textbf{Projection} & \textbf{Ablation sens.} & \textbf{Random-direction rank} & \textbf{Hessian-trace rank} \\
\midrule
\textbf{transformer.adapter}  & \textbf{0.118} \;($2.3\times$ next) & \textbf{17/17 (last)} & \textbf{3/17} \\
core\_block.3.mlp.fc   & 0.051 & 4/17  & 4/17 \\
core\_block.3.mlp.proj & 0.044 & 14/17 & 1/17 \\
core\_block.1.mlp.proj & 0.041 & 15/17 & 2/17 \\
core\_block.2.mlp.fc   & 0.040 & 5/17  & 5/17 \\
\bottomrule
\end{tabular}
\end{table}

\section{Group-Size Dependence}
\label{sec:app-groupsize}
\begin{table}[!htbp]
\centering
\caption{Group-size sweep (simulated INT4, full model, 30 prompts). Grouped settings retain high final-step agreement, while per-channel quantization collapses to 7.6\%. Bold marks the collapsed per-channel endpoint.}
\label{tab:groupsize}
\small
\begin{tabular}{@{}lccc@{}}
\toprule
\textbf{Group size} & \textbf{$N{=}1$} & \textbf{$N{=}8$} & \textbf{$N{=}32$} \\
\midrule
$g{=}32$       & 68.3\% & 90.6\% & 88.4\% \\
$g{=}64$       & 68.8\% & 90.5\% & 87.5\% \\
$g{=}128$      & 68.7\% & 87.7\% & 86.5\% \\
$g{=}256$      & 69.4\% & 86.6\% & 83.9\% \\
\midrule
\textbf{Per-channel}    & 18.8\% &  7.8\% &  \textbf{7.6\%} \\
\bottomrule
\end{tabular}
\end{table}

All grouped configurations maintain 84--88\% top-1 at $N{=}32$, with marginal improvement below $g{=}128$. Per-channel quantization drops to 7.6\%, a $10\times$ gap. The per-channel collapse is consistent with the adapter's $\rho>1$ result in Section~\ref{sec:residual}.

\paragraph{Perturbation scale.}
Input-to-state stability relates bounded disturbances to bounded state responses~\citep{sontag2008iss}. For a locally contractive map with contraction modulus $L<1$, a uniform per-step perturbation $\delta_q$ has a geometric bound proportional to $\delta_q/(1-L)$. The bound explains why a modest increase in $\delta_q$ can move a near-boundary system from bounded error to basin escape.

\section{Cross-Architecture Rank Survey}
\label{sec:app-rank}

\begin{table}[H]
\centering
\caption{Cross-architecture Hessian-rank survey at $10^{-3}\lambda_{\max}$. The bottleneck has the lowest step-0 rank ratio among GPTQ-quantized shared layers. Max rank gain is the largest accumulated-to-step-0 ratio across shared layers. Bold marks checkpoints where one-step GPTQ is worse than RTN.}
\label{tab:rank-survey}
\begingroup
\footnotesize
\setlength{\tabcolsep}{1.8pt}
\renewcommand{\arraystretch}{1.06}
\begin{tabular}{@{}llcccc@{}}
\toprule
\textbf{Architecture} & \textbf{Bottleneck layer} & \textbf{\shortstack{Step-0\\rank}} & \textbf{\shortstack{Max rank\\gain}} & \textbf{\shortstack{1-step\\vs.\ RTN}} & \textbf{\shortstack{Gap\\closed}} \\
\midrule
Ouro 1.4B & layer-0 down\_proj & 0.02\% & 6.2$\times$\textsuperscript{$\dagger$} & \textbf{worse} & 49\% \\
Ouro 2.6B & layer-0 down\_proj & 0.02\% & 5.4$\times$\textsuperscript{$\dagger$} & \textbf{worse} & 17\% \\
Nanbeige4.2-3B-Base & layer-1 down\_proj & 0.02\% & 3.67$\times$ & \textbf{worse} & 58\% \\
R-Llama 1.4B & adapter & 0.02\% & 3.0$\times$ & better & 54\% \\
Parcae 1.3B & core.7 mlp.proj & 0.03\% & 6.9$\times$ & better & 96\% \\
Huginn 3.5B & core.3 mlp.proj & 0.36\% & 7.8$\times$ & \textbf{worse} & 102\% \\
Parcae 140M & core.1 mlp.proj & 0.88\% & 2.1$\times$ & better & 90\% \\
COCONUT 124M & h.2 mlp.c\_proj & 1.40\% & 1.1$\times$ & better & 74\%\textsuperscript{$\S$} \\
LoopFormer 278M & c\_fc (block~0)\textsuperscript{$\ddagger$} & 1.51\% & 32.8$\times$ & \textbf{worse} & 31\% \\
PonderPythia 2.8B & layer-0 dense\_4h\_to\_h & 2.76\% & 2.03$\times$ & better & 43\% \\
\bottomrule
\end{tabular}
\endgroup
\par\smallskip
{\footnotesize\raggedright For Huginn, \texttt{mlp.proj}@3 has the lowest step-0 rank ratio (0.36\%; gain $1.4\times$), while the adapter has the maximum rank gain ($7.8\times$). Gap closed is $(N\text{-step}-RTN)/(Base-RTN)$ and is computed from unrounded scores.

\textsuperscript{$\dagger$} Max rank gain is from a non-bottleneck layer. For both Ouro models, the step-0 Hessian contributes at least 99.5\% of the 4-step trace sum, and adding steps 1--3 does not raise the bottleneck's thresholded rank above 1 at $10^{-3}\lambda_{\max}$. Per-step Hessians at steps 1--3 individually have rank 974--1{,}194 of 5{,}632 (1.4B) and 1{,}010--1{,}335 of 5{,}632 (2.6B).

\textsuperscript{$\ddagger$} The $32.8\times$ maximum belongs to \texttt{attn.c\_proj} in block~0 (rank 50 $\to$ 1{,}640); the GPTQ-relevant c\_fc bottleneck gains $4.1\times$. adaLN modulation layers have rank~1 at step~0 (gain $6.0\times$) and use RTN fallback.

\textsuperscript{$\S$} COCONUT uses in-sample top-1 agreement on 30 prompts (fp32 FakeQuant, 4 latent tokens). One-step GPTQ already closes 70\% of this gap; the trajectory contribution is $+0.8$\,pp top-1 ($+4$\,pp gap-closed).\par}
\end{table}

Parcae-1.3B's bottleneck (core.7 mlp.proj, 0.03\%) drives the per-layer calibration failure even though one-step GPTQ beats RTN at model level; the remaining layers are well-conditioned enough to compensate. Ouro-2.6B's bottleneck has step-0 rank 1/5{,}632 while the other 47 layers are near full rank; a model average would not flag it. Mamba is not included because the SSM experiments measure component divergence rather than a step-0 Hessian-rank sweep.

\begin{table}[!htbp]
\centering
\caption{Step-1 Hessian ranks for the 8 checkpoints with retained per-step traces. For $H_t=X_t^\top X_t$, each count is the number of eigenvalues above $10^{-3}\lambda_{\max}(H_t)$; denominators are input dimensions. Rows use the rank-survey bottleneck. Huginn instead uses its adapter, where its maximum rank gain occurs.}
\label{tab:step1-screen}
\begingroup
\small
\setlength{\tabcolsep}{2.5pt}
\begin{tabular}{@{}llrr@{}}
\toprule
\textbf{Checkpoint} & \textbf{Screened layer} & \textbf{Step 0} & \textbf{Step 1} \\
\midrule
\multicolumn{4}{@{}l}{\emph{1-step GPTQ worse than RTN}} \\
Ouro 1.4B & layer-0 \texttt{down\_proj} & $1/5{,}632$ & \textbf{$974/5{,}632$} \\
Ouro 2.6B & layer-0 \texttt{down\_proj} & $1/5{,}632$ & \textbf{$1{,}010/5{,}632$} \\
Huginn 3.5B & adapter & $43/10{,}560$ & \textbf{$259/10{,}560$} \\
LoopFormer 278M & \texttt{c\_fc} block 0 & $31/2{,}048$ & \textbf{$154/2{,}048$} \\
\midrule
\multicolumn{4}{@{}l}{\emph{1-step GPTQ better than RTN}} \\
COCONUT 124M & h.2 \texttt{mlp.c\_proj} & $43/3{,}072$ & $45/3{,}072$ \\
R-Llama 1.4B & adapter & $1/4{,}096$ & $2/4{,}096$ \\
Parcae 1.3B & core.7 \texttt{mlp.proj} & $2/6{,}144$ & $4/6{,}144$ \\
Parcae 140M & core.1 \texttt{mlp.proj} & $27/3{,}072$ & $21/3{,}072$ \\
\bottomrule
\end{tabular}
\endgroup
\end{table}

A separate Huginn control holds the same 2{,}919 prompt-token positions across steps and gives ranks 361--363 at steps 7, 15, and 31, each slightly above the 32-step accumulated rank of 335.

\section{Frozen-Trajectory Proxy Stability}
\label{sec:app-frozen}

The deployment-weighted reconstruction proxy in Eq.~\ref{eq:deploy} fixes the state trajectory at the bf16 reference. We evaluate fixed per-channel and $g{=}128$ RTN perturbations with the per-step proxy $\mathrm{tr}(\Delta W\, H_t\, \Delta W^\top)$ on bf16 activations, summed over all 17 shared projections (50 GSM8K prompts, Huginn-3.5B). This directly tests whether the proxy's magnitude is stable across the reference rollout.

At $g{=}128$, the per-step proxy varies by only $1.01\times$ across all 32 steps (709.5 at step~0 to 702.5 at step~31), showing that its magnitude is nearly constant along the bf16 rollout at deployment precision. At per-channel, the proxy varies $1.12\times$, with a $1.12\times$ increase from step~0 to step~1, then gradually stabilizes. Since the matched adapter-only per-channel diagnostic has $\rho > 1$ on all 30 prompts, we use the proxy there as a diagnostic rather than an exact loss estimate.

Step~0 accounts for only 3.0\% (per-channel) and 3.2\% ($g{=}128$) of the cumulative deployment proxy $\sum_t \mathrm{tr}(\Delta W\, H_t\, \Delta W^\top)$, close to the uniform expectation of 3.125\%. The proxy magnitude is roughly equal across steps; the calibration gap is directional, not scalar. Only 43 of 10{,}560 adapter input directions exceed the step-0 threshold (0.41\%; Table~\ref{tab:rank}), so single-step GPTQ optimizes rounding over a narrow subspace even though per-step error magnitude is uniform.

\paragraph{Cross-architecture check.} On Ouro-1.4B (4 recurrence steps, 50 WikiText-2 sequences), the $g{=}128$ per-step proxy varies by $1.02\times$ across steps, and step~0 accounts for 25.2\% of the total (uniform expectation: 25.0\%).

\paragraph{Closed-loop INT3 check.}
Recollecting the Hessians from the quantized rollout changes final-step agreement with bf16 by $+0.37$\,pp (prompt-cluster bootstrap 95\% CI, $[-2.31, 3.10]$\,pp). The open-loop arm collects Hessians for all 17 shared linears over the 32-step bf16 rollout and then applies deterministic group-128 INT3 FakeQuant. The closed-loop arm recollects them from that quantized rollout and requantizes the same linears. On 20 calibration-held-out GSM8K training prompts, agreement is 922/1{,}090 (84.59\%) open-loop and 926/1{,}090 (84.95\%) closed-loop.

\section{Jacobian Measurement Details}
\label{sec:app-jacobian}

The spectral radius $\rho$ is the magnitude of the dominant eigenvalue of the recurrence Jacobian $J = \partial s_{t+1}/\partial s_t$. For the matched experiment below, we losslessly promote the fixed states and weights to FP32, compute exact forward-mode Jacobian-vector products, and estimate the largest-magnitude Ritz value with implicitly restarted Arnoldi, without materializing $J$.

\paragraph{Matched spectral-radius protocol.}
For each of 30 prompts, we form the bf16 states after 32 applications of the no-bypass and bypass recurrences from the same initial state. At each operating state, we evaluate four local maps: bf16 and deterministic per-output-row INT4 adapter weights, each without and with the identity bypass. The bf16 reference states and checkpoint/rounded weights are promoted losslessly into the shared FP32 surrogate; all four maps share the same state and FP32 core, with dtype-dependent normalization epsilon pinned to its bf16-effective value.

We use the same two Arnoldi start vectors in every arm and estimate the largest-magnitude Ritz value with $k=4$, $\mathrm{ncv}=64$, tolerance $10^{-5}$, and a 4{,}096-matvec cap, requiring relative Ritz residual below $10^{-3}$ and agreement between starts within $2\times10^{-3}$. All 480 arm/state/start estimates satisfy these criteria; the largest residual is $9.23\times10^{-6}$ and the largest start disagreement is $9.34\times10^{-6}$.

\paragraph{Spectral radius vs.\ spectral norm.}
Arnoldi estimates the eigenvalue of largest magnitude, $\rho(J)=|\lambda_{\max}|$, rather than the largest singular value $\sigma_1(J)=\|J\|_2$. For directions with nonzero projection onto the dominant eigenspace, $\rho$ sets the asymptotic exponential rate, while non-normality can still produce finite-step transient growth. The spectral norm governs worst-case single-step amplification but can overestimate asymptotic behavior when $J$ is non-normal. We use $\rho$ because the experiment concerns repeated application of the same local map; $\sigma_1$ remains a complementary single-step worst-case diagnostic.

\paragraph{Matched spectral result.}
The effect of the identity bypass on the INT4-induced change in $\rho$ depends on the operating point. At the no-bypass bf16 state, mean $\rho$ is $1.064$ for bf16 and $1.145$ after adapter INT4; $\rho>1$ on 9/30 and 30/30 prompts, respectively. The signed INT4 shifts without and with the bypass are $+0.080$ and $+0.101$, giving a difference-in-differences of $+0.021$ (prompt-bootstrap 95\% CI, $[-0.061,0.121]$). At the bypass bf16 state, the corresponding shifts are $-0.340$ and $-0.041$ (difference-in-differences $+0.299$, 95\% CI, $[0.204,0.393]$). The main text therefore uses the no-bypass $\rho>1$ result and the fixed-weight identity-bypass comparison.

\paragraph{Matched identity-path control.}
Let $C$ denote the full four-block residual core and $a_M$ the bf16 or deterministic per-output-row INT4 adapter. We reuse one adapter pair in the full recurrence maps $F_{M,0}(s)=C(a_M([s,e]))$ and $F_{M,1}(s)=C(s+a_M([s,e]))$. For a shared state $s$ and unit direction $v_k$, let $r_{M,i}(s,v_k;\epsilon)=[F_{M,i}(s+\epsilon\|s\|_2v_k)-F_{M,i}(s-\epsilon\|s\|_2v_k)]/(2\epsilon\|s\|_2)$ and $D_i=[\sum_k\|r_{Q,i}-r_{B,i}\|_2^2/\sum_k\|r_{B,i}\|_2^2]^{1/2}$.

On GSM8K training prompts 70--119, 3 recurrent-state initializations shared across conditions, and 32 shared directions, the primary $\epsilon=0.01$ comparison at the no-bypass bf16 step-32 state gives mean $D_0=1.22249$ and mean $D_1=0.713216$. The geometric mean of the prompt-level $D_1/D_0$ ratios corresponds to 41.66\% attenuation (bootstrap 95\% CI, 41.50--41.82\%; all 50 prompt effects share the sign; Holm-adjusted two-sided sign-flip $p=3.0\times10^{-5}$). Across the 150 prompt-initialization units, mean RMS mismatch before dividing by the bf16 response falls from 1.0169 to 0.6366. Under prompt-cluster aggregation, this is 37.40\% attenuation.

The direction holds at both bf16 operating points and all tested radii ($\epsilon\in\{0.005,0.01,0.02\}$; 24.1--60.0\% attenuation). Because the untrained bypass changes the bf16 rollout (10.82\% final-step token agreement), this experiment supports only the local structural claim.

\section{Error Amplification: R-Llama}
\label{sec:app-error-amp}

To measure R-Llama's trajectory displacement on 30 GSM8K prompts, we run the bf16 model for 32 recurrence steps, then start both the bf16 and quantized recurrences from that state and record $\|s^{\text{q}}_t-s^{\text{bf16}}_t\|_2$ after each of 32 further updates.

Under per-channel INT4 (adapter only, core bf16), final mean displacement is 1082. The median per-prompt geometric mean of the 31 consecutive displacement ratios is 1.003. Under grouped INT4 ($g{=}128$), the corresponding values are 300 and 1.006. The ratio is slightly higher at $g{=}128$ because the per-channel curve begins closer to its plateau, leaving less relative growth over the measured window. Median per-update amplification is largest over the first three transitions and then approaches 1, with small fluctuations around it. After 4 updates, both curves are within 1\% of their final displacement. The saturation pattern is consistent with a displaced fixed point $h^*_q \neq h^*_{\text{bf16}}$.

The displacement magnitude tracks quantization precision: 1082 for per-channel INT4 versus 300 at $g{=}128$, a $3.6\times$ reduction that matches the weight-error norm ratio. The same ordering appears on GSM8K: per-channel RTN gives 0\% accuracy, while $g{=}128$ RTN retains 39.5\% (Section~\ref{sec:residual}; Table~\ref{tab:main}).

\section{IIR Filter Experimental Details}
\label{sec:app-iir}

\paragraph{Filter design.}
Four discrete-time IIR filter designs are created using \texttt{scipy.signal}: Butterworth order~4, Chebyshev Type~I order~4, Chebyshev Type~I order~6, and Elliptic order~4. The three order-4 designs use a normalized passband of $[0.2, 0.4]$. The order-6 Chebyshev uses $[0.3, 0.5]$ with 1\,dB ripple to push poles closer to the unit circle ($|z| = 0.984$, analogous to Huginn's $\rho$ near 1). These designs span a range of pole configurations: Butterworth places poles on a circle (maximally flat magnitude response), Chebyshev clusters poles near the unit circle for sharper rolloff, and the Elliptic filter balances passband and stopband ripple.

\paragraph{State-space conversion.}
Each filter's transfer function $(b, a)$ is converted to controllable canonical state-space form $(A, B, C, D)$ via \texttt{scipy.signal.tf2ss}. The matrix $A \in \mathbb{R}^{n \times n}$ is the state-transition (feedback) matrix; $B$, $C$, and $D$ are the input/output matrices. The system dynamics are $x[n{+}1] = Ax[n] + Bu[n]$, $y[n] = Cx[n] + Du[n]$. Poles are the eigenvalues of $A$; the system is stable if and only if all poles lie inside the unit circle ($|z| < 1$).

\paragraph{Quantization methodology.}
Each matrix is quantized independently via uniform signed round-to-nearest at six bit widths: 4, 6, 8, 10, 12, and 16 bits. Two conditions are tested: (1) quantize the transition matrix $A$ only, and (2) quantize $B$, $C$, and $D$ jointly while leaving $A$ fixed. With $q_{\max}=2^{b-1}-1$, the quantizer uses $\Delta=\max|M_{ij}|/q_{\max}$ and $M_q=\Delta\,\mathrm{clip}(\mathrm{round}(M/\Delta),-q_{\max}-1,q_{\max})$. Stability is assessed by computing the eigenvalues of the (quantized or unquantized) $A$ matrix and checking whether $|z|_{\max} > 1$.

\begin{table}[!htbp]
\centering
\caption{IIR stability under quantization. $|z|_{\max}$ is the largest pole magnitude; rows with \textbf{A-quant stable? = No} are unstable ($|z|_{\max} > 1$). Bold rows mark each filter's stability boundary. BCD quantization leaves poles unchanged, while A-quantization destabilizes low-bit regimes.}
\label{tab:iir-full}
\small
\begin{tabular}{@{}lrcccc@{}}
\toprule
\textbf{Filter} & \textbf{Bits} & \textbf{$|z|_{\max}$ (orig)} & \textbf{$|z|_{\max}$ (A-quant)} & \textbf{$|z|_{\max}$ (BCD-quant)} & \textbf{A-quant stable?} \\
\midrule
Butterworth 4 & 4  & 0.915 & 2.593 & 0.915 & No  \\
              & 6  & 0.915 & 1.287 & 0.915 & No  \\
              & 8  & 0.915 & 1.134 & 0.915 & No  \\
              & 10 & 0.915 & \textbf{1.011} & 0.915 & \textbf{No}  \\
              & 12 & 0.915 & \textbf{0.934} & 0.915 & \textbf{Yes} \\
              & 16 & 0.915 & 0.916 & 0.915 & Yes \\
\midrule
Chebyshev I, 4 & 4  & 0.962 & 4.575 & 0.962 & No  \\
               & 6  & 0.962 & 1.441 & 0.962 & No  \\
               & 8  & 0.962 & 1.141 & 0.962 & No  \\
               & 10 & 0.962 & \textbf{1.062} & 0.962 & \textbf{No}  \\
               & 12 & 0.962 & \textbf{0.950} & 0.962 & \textbf{Yes} \\
               & 16 & 0.962 & 0.965 & 0.962 & Yes \\
\midrule
Chebyshev I, 6 & 4  & 0.984 & 5.645 & 0.984 & No  \\
               & 6  & 0.984 & 1.922 & 0.984 & No  \\
               & 8  & 0.984 & 1.602 & 0.984 & No  \\
               & 10 & 0.984 & 1.202 & 0.984 & No  \\
               & 12 & 0.984 & 1.134 & 0.984 & No  \\
               & 16 & 0.984 & \textbf{1.012} & 0.984 & \textbf{No}  \\
\midrule
Elliptic 4     & 4  & 0.970 & 4.569 & 0.970 & No  \\
               & 6  & 0.970 & 1.439 & 0.970 & No  \\
               & 8  & 0.970 & 1.140 & 0.970 & No  \\
               & 10 & 0.970 & \textbf{1.002} & 0.970 & \textbf{No}  \\
               & 12 & 0.970 & \textbf{0.995} & 0.970 & \textbf{Yes} \\
               & 16 & 0.970 & 0.967 & 0.970 & Yes \\
\bottomrule
\end{tabular}
\end{table}

BCD quantization does not perturb $|z|_{\max}$ beyond machine epsilon at any precision, consistent with the fact that $\det(zI - A)$ is independent of $B$, $C$, $D$. A-quantization pushes poles past the unit circle at $\leq 10$ bits for all four designs. The order-6 Chebyshev boundary case (original $|z|_{\max} = 0.984$) remains unstable even at 16 bits, while the Butterworth design (original $|z|_{\max} = 0.915$) stabilizes at 12 bits. This is consistent with the stability-margin interpretation used for Huginn: a system close to the recurrence boundary can cross it under a small perturbation.

\section{Mamba SSM Experimental Details}
\label{sec:app-mamba}

\paragraph{Methodology.}
Each component is tested under two perturbation conditions. (1)~\emph{FakeQuant}: simulated symmetric INT4 quantization (per-tensor for A\_log since $d_\text{state}{=}16 < g{=}128$; per-group $g{=}128$ for 2D weights; biases at full precision). FakeQuant is deterministic and uses each component's actual quantization error norm. (2)~\emph{Gaussian}: norm-matched Gaussian noise at the median INT4 error across components (22.84 for 130M, 15.09 for 2.8B), ensuring every component receives exactly the same perturbation magnitude regardless of parameter count. The Gaussian condition tests whether the component ordering persists under a second perturbation family.
Across the 5 Mamba-130M layers, the \texttt{A\_log} INT4 weight-error norm ranges from 71.1 to 96.0; the Gaussian control uses the common 22.84 norm for every component.

\paragraph{Measurement procedure.}
For each condition (component $\times$ layer), we run the model on 10 prompts of approximately 100 tokens each. At every timestep $t$ of the SSM scan, we record the hidden state $h_t$ under the perturbed model and the hidden state $h_t^{\text{ref}}$ under the unperturbed model, computing the L2 divergence $\|h_t - h_t^{\text{ref}}\|_2$. Five layers are tested (0, 6, 12, 18, 23 for 130M; 0, 16, 32, 48, 63 for 2.8B). FakeQuant runs once (deterministic); Gaussian runs with three random seeds per layer.

\paragraph{Results.}
Across perturbation types and model sizes, divergence from \texttt{A\_log} starts at zero before growing through the scan. With $h_{-1} = 0$, the perturbed $\bar{A}$ term has no state contribution at $t{=}0$. Under INT4 FakeQuant, the measured growth is 1.28--2.00$\times$ at 130M and 1.30--1.71$\times$ at 2.8B. \texttt{in\_proj} and \texttt{x\_proj} produce divergence immediately, but its magnitude and trajectory vary across layers and model sizes. \texttt{out\_proj} produces zero measured SSM-state divergence because it sits after the scanned recurrence.

\paragraph{Connection to Mamba quantization literature.}
Quamba's error bound assumes the transition matrix $A$ is unperturbed. Our experiment perturbs $A$ directly and measures the resulting divergence. QMamba~\citep{qmamba2024} reports component-level sensitivity and attributes it to the long-tailed distribution of $\bar{A}_t$. Bi-Mamba~\citep{bimamba2024} keeps SSM parameters at full precision and binarizes the input/output projections (Section~\ref{sec:related}). These results provide SSM-side evidence for feedback exposure. Mamba's single forward scan already observes the timestep distribution used at deployment, so it is outside our calibration-blindness test.

\paragraph{Identity injection sweep.}
To test whether adding an identity term to the SSM recurrence reduces \texttt{A\_log} sensitivity, we modify the recurrence to $h_t = \alpha \cdot h_{t-1} + \bar{A} \cdot h_{t-1} + \bar{B} \cdot x_t$, where $\alpha = 0$ recovers the original model. We sweep 9 values of $\alpha$ from 0 to 0.5 on Mamba-130M layer 12, with 5 prompts and 3 seeds per condition. Representative rows are shown in Table~\ref{tab:mamba-alpha}. We add Gaussian noise to layer 12's \texttt{A\_log} at $\|\Delta W\|_F = 37.18$, using layer 0's per-row INT4 error norm as the reference magnitude.

\begin{table}[!htbp]
\centering
\caption{Identity injection in Mamba SSM. Adding $\alpha \cdot h_{t-1}$ to the recurrence reduces \texttt{A\_log} sensitivity monotonically from $\alpha \geq 0.15$, consistent with the Huginn identity-bypass result. Intermediate values interpolate smoothly between the shown rows.}
\label{tab:mamba-alpha}
\small
\begin{tabular}{@{}lcc@{}}
\toprule
$\boldsymbol{\alpha}$ & \textbf{Divergence at last $t$} & \textbf{vs.\ $\alpha{=}0$} \\
\midrule
0.00 & 0.127 & baseline \\
0.10 & 0.130 & $+2\%$ \\
\midrule
0.15 & 0.120 & $-6\%$ \\
0.25 & 0.111 & $-13\%$ \\
\textbf{0.50} & \textbf{0.101} & \textbf{$-20\%$} \\
\bottomrule
\end{tabular}
\end{table}

From $\alpha = 0.15$ onward, divergence drops monotonically, reaching a 20\% reduction at $\alpha = 0.5$, with diminishing returns beyond $\alpha \approx 0.25$. In near-marginal Huginn, the matched identity-bypass control reduces the discrepancy between bf16 and INT4 responses to small state perturbations by 41.7\%.

\section{Eigenvalue Spectrum Analysis}
\label{sec:app-eigenvalue}

\begin{figure}[H]
\centering
\includegraphics[width=\linewidth]{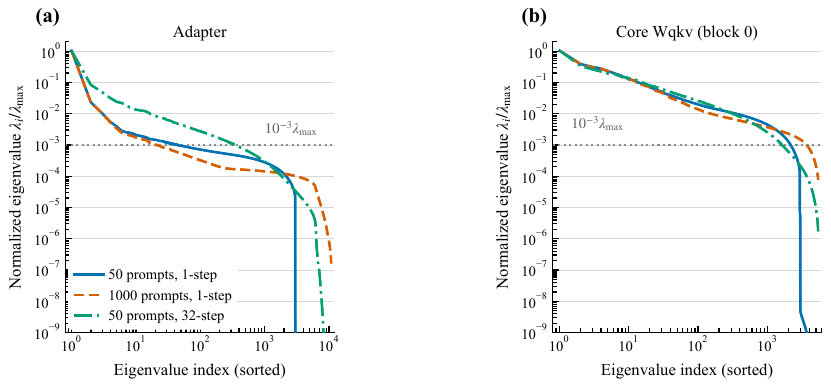}
\caption{Eigenvalue spectrum of the row-normalized calibration Hessian $\widetilde H = 2X^\top X/m$. (a)~Adapter ($d{=}10{,}560$): the spectrum concentrates in 2--3 dominant eigenvalues at step~0, with a sharp drop exceeding 2 orders of magnitude by index 5. Increasing prompts from 50 to 1{,}000 (dashed) \emph{deflates} near-threshold eigenvalues below the $10^{-3}\lambda_{\max}$ cutoff, while $\lambda_{\max}$ is stable ($1.00\times$ ratio). Accumulating 32 steps (dash-dot) lifts the bulk spectrum, populating 335 directions above threshold. (b)~Core Wqkv block~0 ($d{=}5{,}280$): the spectrum decays gradually with no sharp cutoff, and 2{,}111 eigenvalues exceed the threshold at 50 prompts, $49\times$ more than the adapter. The core layer has a much broader step-0 spectrum than the adapter, while 32-step accumulation substantially broadens the adapter spectrum.}
\label{fig:eigenvalue-spectrum}
\end{figure}

Figure~\ref{fig:eigenvalue-spectrum} plots the full eigenvalue spectrum of the row-normalized calibration Hessian $\widetilde H = 2X^\top X/m$, where $m$ is the number of activation rows, for the adapter and a representative core layer (Wqkv, block~0) under three conditions: 50 prompts at step~0, 1{,}000 prompts at step~0, and 50 prompts accumulated across 32 steps.

Table~\ref{tab:eigenvalue-sweep} reports the adapter Hessian rank across multiple thresholds, plus effective rank (erank) and stable rank, under the three calibration conditions. One representative core layer is included for comparison.

\begin{table}[!htbp]
\centering
\caption{Hessian rank across thresholds $\tau$ relative to $\lambda_{\max}$, with effective rank (erank) and stable rank. Core Wqkv from block~0 provides the core-layer comparison. The focal $\tau{=}10^{-3}$ column is bold.}
\label{tab:eigenvalue-sweep}
\small
\setlength{\tabcolsep}{2.4pt}
\begin{tabular}{@{}lrrrrrr@{}}
\toprule
\textbf{Condition} & $\boldsymbol{10^{-1}}$ & $\boldsymbol{10^{-2}}$ & $\boldsymbol{10^{-3}}$ & $\boldsymbol{10^{-4}}$ & $\boldsymbol{10^{-5}}$ & $\boldsymbol{10^{-6}}$ \\
\midrule
\multicolumn{7}{@{}l}{\emph{Adapter} ($d = 10{,}560$)} \\
\quad 50p, 1-step   & 1  & 2  & \textbf{43}  & 2{,}120  & 2{,}919 & 2{,}919 \\
\quad 1000p, 1-step & 1  & 2  & \textbf{20}  & 3{,}047  & 7{,}462 & 9{,}621 \\
\quad 50p, 32-step  & 1  & 16 & \textbf{335} & 1{,}795  & 4{,}796 & 6{,}213 \\
\midrule
\multicolumn{7}{@{}l}{\emph{Core Wqkv@0} ($d = 5{,}280$)} \\
\quad 50p, 1-step   & 12 & 323 & \textbf{2{,}111} & 2{,}870 & 2{,}870 & 2{,}919 \\
\quad 1000p, 1-step & 12 & 142 & \textbf{3{,}588} & 5{,}227 & 5{,}280 & 5{,}280 \\
\quad 50p, 32-step  & 15 & 297 & \textbf{1{,}587} & 3{,}611 & 4{,}890 & 5{,}280 \\
\bottomrule
\end{tabular}
\par\smallskip
\begin{tabular}{@{}lrr@{}}
\toprule
\textbf{Condition} & \textbf{erank} & \textbf{Stable rank} \\
\midrule
\multicolumn{3}{@{}l}{\emph{Adapter}} \\
\quad 50p, 1-step & 51 & 1.8 \\
\quad 1000p, 1-step & 80 & 1.8 \\
\quad 50p, 32-step & 137 & 2.7 \\
\midrule
\multicolumn{3}{@{}l}{\emph{Core Wqkv@0}} \\
\quad 50p, 1-step & 912 & 16.6 \\
\quad 1000p, 1-step & 1{,}548 & 16.8 \\
\quad 50p, 32-step & 705 & 15.1 \\
\bottomrule
\end{tabular}
\par\smallskip
{\footnotesize\raggedright For $p_i=\lambda_i/\sum_j\lambda_j$, effective rank is $\exp(-\sum_i p_i\log p_i)$, and stable rank is $\mathrm{tr}(H)/\lambda_{\max}$. Counts at $\tau\leq10^{-4}$ for the 50-prompt 1-step condition are capped by the sample-limited rank of $H$ ($\sim2{,}919$).\par}
\end{table}

The adapter's maximum eigenvalue changes by $<0.3\%$ between 50 and 1{,}000 prompts, so the rank decrease from 43 to 20 is not a threshold artifact from a growing $\lambda_{\max}$. Instead, near-threshold eigenvalues shrink below the cutoff as the sample count rises. Core layers behave differently. Their rank increases from 2{,}111 to 3{,}588 because the step-0 inputs occupy more directions in input space. Threshold-free measures preserve the same ordering. Adapter erank is $51$ vs.\ $912$ for core Wqkv ($17.9\times$ gap), and stable rank is $1.8$ vs.\ $16.6$ ($9.2\times$ gap). Accumulating 32 steps lifts the adapter rank by $7.8\times$ (43 $\to$ 335) as recurrence steps 1--31 raise more input directions above the threshold.

\paragraph{Step-0 and late-step subspaces.}
On GSM8K training prompts 70--119, we decompose the adapter inputs at steps~0 and~31 under 2 recurrent-state initialization seeds. After restricting the comparison to the changing 5{,}280-dimensional recurrent-state half, we form uncentered second moments. The mean squared cosine of the principal angles between their top-32 eigenspaces at steps~0 and~31 is 0.571\% and 0.631\% for the two seeds. The fixed prompt-embedding half is identical across steps and is excluded. Comparing the seeds directly at $k{=}64$, mean squared-cosine overlap is 99.71\% for the step-31 subspaces but only 1.25\% for the step-0 subspaces. The step-31 subspace is reproducible across seeds, while the step-0 subspace is neither aligned with it nor stable across initializations.

\paragraph{Adapter output magnitude.}
Adapter output norms support the identity-path interpretation in Section~\ref{sec:residual}: $\|\operatorname{Adapter}(\operatorname{cat}(s_t,e))\|_2 / \|s_t\|_2 = 0.415 \pm 0.044$ at steady state (steps 1--31, 50 prompts). The adapter output is non-negligible relative to $s_t$. Adding a residual path $s_t + \text{Adapter}(\text{cat}(s_t, e))$ gives the state an identity channel around that term.

\section{Step-Window Ablation}
\label{sec:app-step-window}

Algorithm~\ref{alg:trajectory} accumulates $H = \sum_{t=0}^{N-1} H_t$ over all recurrence steps. We test whether excluding step~0 improves GPTQ quality on 6 architectures (Table~\ref{tab:step-window}).

\begin{table}[!htbp]
\centering
\caption{Step-0 exclusion ablation. WikiText-2 PPL ($\downarrow$) for grouped-INT4 $g{=}128$. $\Delta$ is skip-0 minus $N$-step; negative means skip-0 is better.}
\label{tab:step-window}
\small
\begin{tabular}{@{}lcrrrr@{}}
\toprule
\textbf{Model} & $N$ & \textbf{RTN} & \textbf{$N$-step} & \textbf{Skip-0} & $\boldsymbol{\Delta}$ \\
\midrule
Ouro 2.6B & 4 & 11.94 & 11.62 & \textbf{11.10} & \textbf{$-$0.52} \\
Parcae 1.3B & 8 & 51.30 & 20.84 & \textbf{20.58} & \textbf{$-$0.26} \\
Huginn 3.5B & 32 & 14.98 & \textbf{14.89} & 14.92 & $+$0.03 \\
R-Llama 1.4B & 32 & 34.02 & \textbf{33.46} & 33.65 & $+$0.19 \\
LoopFormer 278M & 8 & 35.10 & \textbf{34.08} & 34.33 & $+$0.25 \\
Ouro 1.4B & 4 & 14.38 & \textbf{12.97} & 13.78 & $+$0.81 \\
\bottomrule
\end{tabular}
\end{table}

Step~0 activations come from prompt embeddings, before recurrence has shaped the state. Table~\ref{tab:step-window} shows the effect of excluding them is architecture-dependent. Ouro-2.6B ($-0.52$ PPL) and Parcae-1.3B ($-0.26$) improve under skip-0. Ouro-2.6B's bottleneck has step-0 trace $7{,}249\times$ larger than step-1, so the accumulated Hessian is dominated by the step-0 subspace (rank~1). The Hessian assigns near-zero cost to the 1{,}010--1{,}335 directions that later steps contribute, so column compensation can push rounding error into those directions while reducing step-0 reconstruction error (Section~\ref{sec:method}). Excluding step-0 forces the solver to penalize error in those directions. LoopFormer ($+0.25$) and Ouro-1.4B ($+0.81$) degrade: step~0 contributes directions that the remaining steps do not fully cover (Table~\ref{tab:rank-survey}). Huginn ($+0.03$) and R-Llama ($+0.19$) do not benefit.

For Ouro-2.6B, downweighting rather than excluding step~0 confirms the pattern: $H = \alpha H_0 + (1{-}\alpha)H_{\text{rest}}$ degrades monotonically as $\alpha$ grows from 0 to 0.25, with PPL 11.10, 11.28, 11.36, 11.47, 11.60. For Ouro-1.4B the optimum is $\alpha{=}0.15$ (PPL~12.93 vs.\ 12.97 at $\alpha{=}1/N$). Skip-0 is a per-checkpoint tuning knob; full accumulation is the default.

\paragraph{Limits of the per-layer proxy.} Both the deployment proxy (Eq.~\ref{eq:deploy}) and GPTQ treat each layer independently. At step $t$, though, activations $X_t$ already carry rounding errors from steps $0$ through $t{-}1$. The frozen-trajectory proxy fixes $X_t$ at the bf16 reference and drops these cross-step terms. On both Ouro scales, skip-0 reduces per-layer $\mathrm{tr}(\Delta W\, H_3\, \Delta W^\top)$ nearly everywhere (335/336 layers at 2.6B, 168/168 at 1.4B). PPL improves only at 2.6B. The per-layer sum ignores this coupling, so small proxy gains need not translate into lower model-level perplexity.

\section{Experimental Setup}
\label{sec:app-setup}

\paragraph{Benchmarks.}
GSM8K~\citep{gsm8k} (8-shot chain-of-thought, $n{=}1{,}319$) is the primary generation benchmark for Huginn/R-Llama. Strict match counts exact normalized numeric answers; flex match accepts extractor-tolerant equivalent numeric answers, and intervals are Wilson 95\% confidence intervals where shown. WikiText-2~\citep{merity2017pointer} perplexity is the primary metric for Ouro, LoopFormer, Parcae, PonderPythia, and Nanbeige and is evaluated on the standard test split. LAMBADA~\citep{paperno2016lambada} accuracy is reported as supporting evidence for Parcae. ARC-Challenge~\citep{clark2018think} (0-shot, 1{,}172 problems) and HellaSwag~\citep{zellers2019hellaswag} (0-shot, 10{,}042 problems) use log-likelihood scoring. Per-token fidelity is measured as top-1 agreement with bf16 and KL divergence across all recurrence steps. Benchmark evaluations use the \texttt{lm-eval-harness} framework~\citep{gao2024lmeval}.

\begin{table}[!htbp]
\centering
\caption{Wilson 95\% confidence intervals for the main GSM8K rows evaluated on all 1{,}319 test examples, computed from the exact underlying counts.}
\label{tab:gsm8k-ci}
\small
\begin{tabular}{@{}llcc@{}}
\toprule
\textbf{Model} & \textbf{Config} & \textbf{Strict (\%)} & \textbf{Wilson 95\% CI} \\
\midrule
Huginn 3.5B & bf16 & 34.9 & [32.4, 37.5] \\
Huginn 3.5B & RTN & 27.9 & [25.5, 30.4] \\
Huginn 3.5B & 1-step GPTQ & 26.9 & [24.6, 29.4] \\
Huginn 3.5B & \textbf{$N$-step GPTQ} & \textbf{35.0} & [32.5, 37.6] \\
\midrule
R-Llama 1.4B & bf16 & 50.2 & [47.5, 52.9] \\
R-Llama 1.4B & RTN & 39.5 & [36.9, 42.2] \\
R-Llama 1.4B & 1-step GPTQ & 40.3 & [37.6, 42.9] \\
R-Llama 1.4B & \textbf{$N$-step GPTQ} & \textbf{45.3} & [42.6, 48.0] \\
\bottomrule
\end{tabular}
\end{table}

\paragraph{Quantization and hardware.}
All experiments run on a single NVIDIA A100 (CUDA 12.4, PyTorch 2.6). Main-table quantization uses deterministic dequantized grouped-INT4 FakeQuant ($g{=}128$), isolating calibration effects from integer-kernel behavior. Huginn/R-Llama use a local implementation of GPTQ's column-update rule~\citep{frantar2022gptq}; Ouro, LoopFormer, Parcae, PonderPythia, and Nanbeige use architecture-specific implementations of the same rule. Parcae quantizes \texttt{core\_block} only.

PonderPythia uses 3 pondering updates plus a final shared-stack prediction pass and quantizes all 128 shared-stack projections plus its untied output projection (129 total). Nanbeige applies its 22-layer stack twice and quantizes 154 shared-stack projections plus the output projection (155 total; 3.66B weights); in both GPTQ arms the output projection uses the same RTN weights. Input embeddings remain bf16 in both models. The Marlin W4A16 kernel~\citep{marlin2024} is used for the real-kernel speed measurement (Section~\ref{sec:system}). INT8 rows, where referenced in appendix ablations, use simulated per-channel symmetric quantization (FakeQuant with fp32 arithmetic). Mamba experiments use deterministic INT4 FakeQuant plus norm-matched Gaussian controls. On Huginn, logged Hessian collection took approximately 15 minutes for one-step calibration and 25 minutes for 32-step trajectory calibration on one A100; the calibration change adds no inference-time cost.

\begin{table}[!htbp]
\centering
\caption{Calibration and solver details for the recovery rows. All rows use deterministic FakeQuant with $g{=}128$, no act-order, and unchanged inference graphs within each row.}
\label{tab:calibration-details}
\footnotesize
\setlength{\tabcolsep}{2.5pt}
\begin{tabular}{@{}>{\raggedright\arraybackslash}p{0.17\linewidth}>{\raggedright\arraybackslash}p{0.27\linewidth}>{\raggedright\arraybackslash}p{0.20\linewidth}>{\raggedright\arraybackslash}p{0.29\linewidth}@{}}
\toprule
\textbf{Rows} & \textbf{Calibration data} & \textbf{Hessian collection} & \textbf{Solver} \\
\midrule
Huginn/R-Llama & First 50 GSM8K training questions & 1 or 32 recurrence steps & GPTQ, 1\% damping for core and 10\% for adapter \\
Ouro & WikiText-2 train, 128$\times$2048 tokens & 1 or 4 recurrence steps & Column-wise GPTQ, 0.01 damping \\
LoopFormer & WikiText-2 train, 128$\times$1024 tokens & 1 or 8 loop iterations & Column-wise GPTQ, 0.01 damping \\
Parcae & WikiText-2 train, 128$\times$1024 tokens & 1 or 8 core iterations & Column-wise GPTQ, 0.01 damping \\
PonderPythia & WikiText-2 train, 128$\times$2048 tokens & First pass or all 4 shared-stack passes & Column-wise GPTQ, 0.01 damping \\
Nanbeige & WikiText-2 train, 128$\times$2048 tokens & First pass or both shared-stack passes & Column-wise GPTQ, 0.01 damping \\
\bottomrule
\end{tabular}
\end{table}

\paragraph{Third-party assets.}
All third-party checkpoints, datasets, and tools are accessed from upstream sources; none are redistributed with this submission. The table records the license metadata stated by those sources and marks checkpoints whose cards state no license terms. Links point to exact revisions when a pinned revision is available; the table displays shortened hashes.

\begin{table}[H]
\centering
\caption{Third-party asset licenses. No third-party checkpoints, datasets, or tooling are redistributed with this submission.}
\label{tab:asset-status}
\small
\setlength{\tabcolsep}{2pt}
\begin{tabular}{@{}>{\raggedright\arraybackslash}p{0.68\linewidth}>{\raggedright\arraybackslash}p{0.27\linewidth}@{}}
\toprule
\textbf{Asset / source} & \textbf{License} \\
\midrule
\multicolumn{2}{@{}l}{\textbf{Checkpoints}} \\
\addlinespace[1pt]
\href{https://huggingface.co/tomg-group-umd/huginn-0125/tree/0f0fa0ba5cf9d3daad2db05811412af74d111904}{Huginn-3.5B}~(\texttt{0f0fa0ba});
\href{https://huggingface.co/smcleish/Recurrent-Llama-3.2-train-recurrence-32/tree/a5f6f126e9d9302445346844f1d7d29d111a06eb}{R-Llama-1.4B}~(\texttt{a5f6f126});
\href{https://huggingface.co/ByteDance/Ouro-1.4B/tree/574fa66cb8bf5abdc979642d01cf2b79b16bfab1}{Ouro-1.4B}~(\texttt{574fa66c});
\href{https://huggingface.co/ByteDance/Ouro-2.6B/tree/1ed04250da1a9936042725d302e81c8fa2ab5abd}{Ouro-2.6B}~(\texttt{1ed04250}) & Apache-2.0 (pinned checkpoint cards). \\
\href{https://huggingface.co/zeng123/PonderingPythia-2.8B/tree/5bd111c10a7283e4afd672cca15a54666a41d0f6}{PonderPythia-2.8B}~(\texttt{5bd111c1});
\href{https://huggingface.co/Nanbeige/Nanbeige4.2-3B-Base/tree/4a38e817c14b9f2c4b69b1e05cbda90395872f6b}{Nanbeige4.2-3B-Base}~(\texttt{4a38e817}) & Apache-2.0 (pinned checkpoint cards). \\
\href{https://huggingface.co/state-spaces/mamba-130m-hf}{Mamba-130M};
\href{https://huggingface.co/state-spaces/mamba-2.8b-hf}{Mamba-2.8B} & Checkpoint terms not stated; code Apache-2.0. \\
\href{https://huggingface.co/armenjeddi/LoopFormer-3block-8iterations-FineWeb300K/tree/e7c87640da4b7a2315b720b3f9c1c068432f4663}{LoopFormer-278M}~(\texttt{e7c87640});
\href{https://huggingface.co/SandyResearch/parcae-140m/tree/236739ce93ba1afc1b3f172fb4a509f4500d2852}{Parcae-140M}~(\texttt{236739ce});
\href{https://huggingface.co/SandyResearch/parcae-1.3b/tree/517f8e6add927e3d73e15beec13a5b500537b9cc}{Parcae-1.3B}~(\texttt{517f8e6a}) & Checkpoint terms not stated; code MIT. \\
\href{https://huggingface.co/Onlydrinkwater/gpt2-coconut-checkpoint14}{COCONUT-GPT2-124M} & Checkpoint terms not stated; code MIT. \\
\addlinespace[2pt]
\multicolumn{2}{@{}l}{\textbf{Benchmarks}} \\
\addlinespace[1pt]
GSM8K & MIT. \\
WikiText-2 (Salesforce; EleutherAI document-level) & CC BY-SA 3.0; Salesforce source card also lists GFDL. \\
LAMBADA & Modified MIT. \\
ARC-Challenge & CC BY-SA 4.0. \\
HellaSwag & MIT. \\
\addlinespace[2pt]
\multicolumn{2}{@{}l}{\textbf{Tooling}} \\
\addlinespace[1pt]
\texttt{lm-eval-harness} & MIT. \\
IST-DASLab GPTQ; Marlin & Apache-2.0. \\
PyTorch & BSD-3-Clause. \\
HuggingFace Transformers & Apache-2.0. \\
\bottomrule
\end{tabular}
\end{table}

\end{document}